%% file: main.tex
\documentclass[11pt, a4paper]{googledeepmind}

\pdftrailerid{redacted}

\makeatletter
\providecommand\bibentry[1]{\nocite{#1}\fullcite{#1}}
\makeatother

\usepackage[
  style=authoryear-comp,
  sorting=ynt,
  natbib=true,
  backend=bibtex,
  maxcitenames=2,
  maxbibnames=99,
  uniquelist=false,
  uniquename=false,
  giveninits=true,
  dashed=false
]{biblatex}
\let\cite\textcite

\DeclareFieldFormat{citehyperref}{%
  \DeclareFieldAlias{bibhyperref}{noformat}%
  \bibhyperref{#1}}
\DeclareFieldFormat{textcitehyperref}{%
  \DeclareFieldAlias{bibhyperref}{noformat}%
  \bibhyperref{%
    #1%
    \ifbool{cbx:parens}
      {\bibcloseparen\global\boolfalse{cbx:parens}}
      {}}}
\savebibmacro{cite}
\savebibmacro{textcite}
\renewbibmacro*{cite}{%
  \printtext[citehyperref]{%
    \restorebibmacro{cite}%
    \usebibmacro{cite}}}
\renewbibmacro*{textcite}{%
  \ifboolexpr{
    ( not test {\iffieldundef{prenote}} and
      test {\ifnumequal{\value{citecount}}{1}} )
    or
    ( not test {\iffieldundef{postnote}} and
      test {\ifnumequal{\value{citecount}}{\value{citetotal}}} )
  }
    {\DeclareFieldAlias{textcitehyperref}{noformat}}
    {}%
  \printtext[textcitehyperref]{%
    \restorebibmacro{textcite}%
    \usebibmacro{textcite}}}

\usepackage[utf8]{inputenc}
\usepackage[T1]{fontenc}
\AtBeginDocument{%
  \hypersetup{
    linkcolor=RoyalBlue!85!black,
    citecolor=RoyalBlue!85!black,
    urlcolor=RoyalBlue!85!black,
  }%
}
\usepackage{url}
\usepackage{booktabs}
\usepackage{amsfonts,amsmath,amssymb}
\usepackage{nicefrac}
\usepackage{microtype}
\usepackage{graphicx}
\usepackage{subcaption}
\usepackage{multirow}
\usepackage{makecell}
\usepackage{enumitem}
\usepackage{pifont}
\usepackage{float}
\usepackage{placeins}
\usepackage[table,dvipsnames]{xcolor}
\usepackage{threeparttable}
\usepackage{colortbl}
\usepackage{adjustbox}
\usepackage{caption}
\usepackage[nameinlink]{cleveref}
\usepackage{gdm-colors}

\usepackage[breakable,skins]{tcolorbox}
\usepackage{listings}

\definecolor{AccentBlue}{HTML}{2563EB}
\definecolor{AccentTeal}{HTML}{0D9488}
\definecolor{SoftGray}{HTML}{F8F9FA}
\definecolor{MedGray}{HTML}{E5E7EB}
\definecolor{DarkText}{HTML}{1F2937}
\definecolor{RowHighlight}{HTML}{EFF6FF}
\definecolor{PromptBg}{HTML}{F9FAFB}
\definecolor{PromptFrame}{HTML}{D1D5DB}
\definecolor{PromptTitle}{HTML}{1E40AF}

\lstdefinestyle{promptstyle}{
  basicstyle=\ttfamily\scriptsize\color{DarkText},
  breaklines=true,
  breakatwhitespace=false,
  columns=fullflexible,
  keepspaces=true,
  escapeinside={(*@}{@*)},
  aboveskip=0pt,
  belowskip=0pt,
}

\newtcolorbox{promptbox}[1][]{
  enhanced,
  breakable,
  colback=PromptBg,
  colframe=PromptFrame,
  coltitle=white,
  colbacktitle=PromptTitle,
  fonttitle=\small\bfseries\sffamily,
  boxrule=0.5pt,
  arc=2pt,
  left=6pt, right=6pt, top=4pt, bottom=4pt,
  toptitle=2pt, bottomtitle=2pt,
  title={#1},
}

\newcommand{\second}[1]{\underline{#1}}
\newcommand{\up}[1]{\,{\scriptsize\textcolor{AccentTeal}{(+#1)}}}
\newcommand{\down}[1]{\,{\scriptsize\textcolor{red}{(\textminus#1)}}}

\title{GoLongRL: Capability-Oriented Long Context\\[2pt]
Reinforcement Learning with Multitask Alignment}

\renewcommand{\today}{}

\author{%
  \bfseries
  Minxuan Lv\textsuperscript{*,1}\enspace
  Tiehua Mei\textsuperscript{*,1}\enspace
  Tanlong Du\textsuperscript{*,1}\enspace
  Junmin Chen\textsuperscript{1}\enspace
  Zhenpeng Su\textsuperscript{$\dagger$,2}\enspace\\
  \bfseries
  \fontsize{9}{13}\selectfont
  Ziyang Chen\textsuperscript{2}\enspace
  Ziqi Wang\textsuperscript{1}\enspace
  Zhennan Wu\textsuperscript{1}\enspace
  Ruotong Pan\textsuperscript{1}\enspace
  Jian Liang\textsuperscript{1}\enspace
  Ruiming Tang\textsuperscript{$\dagger$,1}\enspace
  Han Li \textsuperscript{$\dagger$,1}\\
  \fontsize{8}{13}\selectfont
  \normalfont\textsuperscript{1} Kuaishou Technology\qquad
  \textsuperscript{2} University of Chinese Academy of Sciences\\
  \normalfont\textsuperscript{*}Equal contribution\qquad
  \textsuperscript{$\dagger$}Corresponding author
}

\begin{abstract}
We present \textbf{GoLongRL}, a fully open-source, capability-oriented post-training recipe for long-context reinforcement learning with verifiable rewards (RLVR). Existing long-context RL methods often treat data construction as a matter of designing increasingly complex retrieval paths, leading to homogeneous task coverage and reward formulations that inadequately reflect practical long-context requirements. Our work offers two contributions. \textbf{(1) Capability-oriented data construction with full open release.} We openly release a dataset of 23K RLVR samples, the complete construction pipeline, and all training code. Guided by a taxonomy of long-context capabilities, the dataset spans 9 task types, each paired with its natural evaluation metric. It comprises curated open-source samples from established corpora and synthetic samples whose QA pairs are generated from real source documents such as books, academic papers, and multi-turn dialogues. Under the same vanilla GRPO setup, our dataset alone outperforms the closed-source QwenLong-L1.5 dataset. Moreover, our Qwen3-30B-A3B model trained on this data delivers long-context performance comparable to DeepSeek-R1-0528 and Qwen3-235B-A22B-Thinking-2507, suggesting that broader coverage and greater reward diversity substantially benefit long-context capability improvement. \textbf{(2) TMN-Reweight for heterogeneous multitask optimization.} To address optimization challenges from heterogeneous rewards, we propose TMN-Reweight, which combines task-level mean normalization for cross-task reward scale alignment with difficulty-adaptive weighting for more reliable advantage estimation. TMN-Reweight further improves average performance over vanilla GRPO, with general capabilities preserved or improved across reported evaluations. 

All resources, including the dataset and training code, are publicly available at \url{https://github.com/xiaoxuanNLP/GoLongRL}.
\end{abstract}

\begin{document}
\maketitle

\input{sections/abstract}
\clearpage
\input{sections/introduction}
\input{sections/related_work}
\input{sections/data}
\input{sections/algorithm}
\input{sections/experiments}
\input{sections/conclusion}

\printbibliography

\appendix
\input{sections/appendix}

\end{document}

%% file: sections/abstract.tex
\begin{figure}[H]
    \centering
    \includegraphics[width=\textwidth]{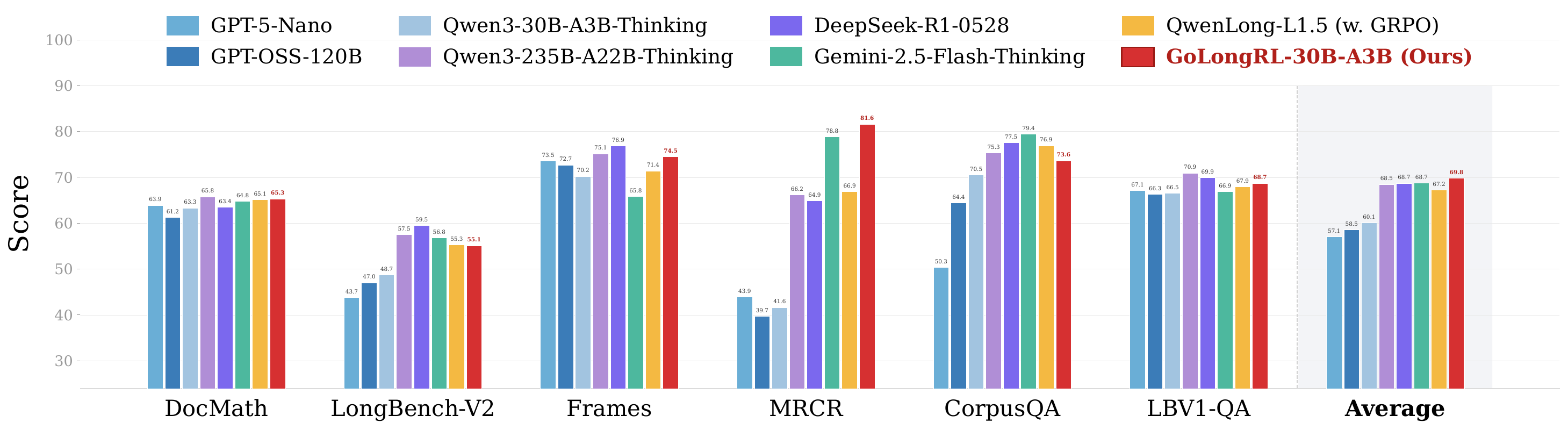}
    \caption{Overall performance comparison on long-context benchmarks.}
    \label{fig:benchmark_overall_30b}
\end{figure}

%% file: sections/introduction.tex
\section{Introduction}
\label{sec:intro}

As large language models (LLMs) are increasingly deployed in practical domains, long-context reasoning has emerged as an important capability for advanced AI systems~\citep{achiam2023gpt, liu2024lost}. Tasks such as multi-document analysis, deep research, retrieval-augmented generation (RAG), and agentic workflows often require coherent understanding over tens or hundreds of thousands of tokens~\citep{bai2024longbench}. Although current LLM foundations have rapidly expanded their context windows through pretraining and mid-training~\citep{qwen3, deepseekv32, chu2026kwai}, effective use of long contexts during post-training remains a significant bottleneck. Recent studies suggest that reinforcement learning (RL) can further improve how effectively these extended contexts are utilized~\citep{deepseekr1, loongrl, longrlvr}.

Existing RL-based long-context methods share a common limitation in both data design and optimization. On the data side, training data is frequently constructed around complex retrieval paths, including UUID chain tracking, chunk-based QA, and similar patterns. While such datasets provide useful supervision, they tend to produce narrow task coverage, artificial difficulty structures, and overly uniform reward design. Important capabilities such as summarization, ranking, aggregation, and structured reasoning consequently receive limited direct supervision. On the algorithm side, standard GRPO can introduce two optimization issues when applied to heterogeneous tasks. Per-prompt normalization may distort advantage estimates across prompts of varying difficulty, and reward metrics such as EM, F1, NDCG, and ROUGE-L exhibit different variance profiles, causing high-variance tasks to contribute disproportionate gradients.

% We address these challenges through a capability-oriented long-context RL framework. On the data side, inspired by the task taxonomy of LongBench Pro~\citep{longbenchpro}, we define 9 task types that cover core capabilities for long-context understanding and construct 23K RLVR samples across these tasks. Each task is paired with its natural evaluation metric as the reward function, rather than being collapsed into a single indicator. The dataset draws from two complementary pools, with roughly 14K curated open-source samples adapted from established corpora and roughly 9K synthetic samples constructed from real source material, including books, academic papers, and multi-turn dialogues. The synthetic data is further processed through a two-stage quality control pipeline that first calibrates sample difficulty via multi-scale pass rate bucketing and then refines the remaining samples by identifying and correcting recurring failure modes such as ambiguous questions and incorrect labels. On the algorithm side, we propose TMN-Reweight, or Task-level Mean Normalization with difficulty-adaptive Reweighting. Task-level normalization replaces the per-prompt $\sigma_u$ with a task-level root mean square standard deviation, reducing cross-task scale differences without erasing individual prompt difficulty. Difficulty-adaptive weighting then uses a smoothed pass rate to emphasize informative hard positive samples while reducing redundant signals from easy cases.

We address these challenges through a capability-oriented long-context RL framework. On the data side, inspired by the task taxonomy of LongBench Pro~\citep{longbenchpro}, we define 9 task types that cover core capabilities for long-context understanding and construct 23K RLVR samples across these tasks. Each task is paired with its natural evaluation metric as the reward function, rather than being collapsed into a single indicator. The dataset draws from two complementary pools, with roughly 14K curated open-source samples adapted from established long-context corpora with existing annotations, and roughly 9K synthetic samples whose QA pairs are generated from real source documents including books, academic papers, and multi-turn dialogues. The entire dataset is produced through a four-phase pipeline (Section~\ref{sec:data_pipeline}) that covers source collection, task-oriented filtering, sample construction with multi-stage quality control, and iterative refinement guided by benchmark diagnostics. On the algorithm side, we propose TMN-Reweight, short for Task-level Mean Normalization with difficulty-adaptive Reweighting. Task-level normalization replaces the per-prompt $\sigma_u$ with a task-level root mean square standard deviation, reducing cross-task scale differences without erasing individual prompt difficulty. Difficulty-adaptive weighting then uses a smoothed pass rate to emphasize informative hard positive samples while reducing redundant signals from easy cases.

Under the same GRPO setting, training on our capability-oriented dataset improves the long-context average from 53.0 to 62.2 on Qwen3-4B-Thinking and from 60.1 to 69.8 on Qwen3-30B-A3B, outperforming QwenLong-L1.5 trained with GRPO at both scales. Extensive ablation studies on the 4B model further show that TMN-Reweight raises the average to 63.0, surpassing QwenLong-L1.5. General capabilities remain stable or improve on the reported evaluations, including MMLU-Pro, AIME24/25, and GPQA. Our contributions can be summarized as follows.
\begin{enumerate}[leftmargin=*,itemsep=2pt]
    % \item \textbf{Data.} We construct a capability oriented long-context RLVR dataset with 23K samples across 9 task types, accompanied by a data construction pipeline that curates long-context training samples from real documents and open-source corpora. All code and datasets have been open-sourced.
    \item \textbf{Data.} A capability-oriented long-context RLVR dataset of 23K samples across 9 task types with heterogeneous reward functions. Vanilla GRPO trained on our data outperforms QwenLong-L1.5 trained with GRPO at both the 4B and 30B scales, suggesting data coverage and reward diversity are primary bottlenecks in long-context RL.
    % \item \textbf{Algorithm.} We introduce TMN-Reweight, a method that jointly mitigates cross-task reward scale inconsistency and difficulty bias in multitask RL, yielding consistent gains at the 4B scale.
    \item \textbf{Algorithm.} TMN-Reweight jointly mitigates cross-task reward scale inconsistency and difficulty-induced advantage bias, yielding consistent gains over vanilla GRPO at both scales.
    % \item \textbf{Empirical findings.} We show that long-context RL with diverse capability oriented data can match or exceed dedicated long-context methods while preserving general capabilities and exhibiting transfer to related capabilities such as dialogue memory.
    % \item  \textbf{Generalization.} Long-context RL with capability-oriented data preserves or improves general reasoning and memory capabilities, including transfer to dialogue memory tasks not seen during training.
    \item \textbf{Generalization.} Long-context RL with capability-oriented data preserves or improves general reasoning and memory capabilities, including transfer to agentic memory and long-term dialogue memory benchmarks not seen during training.
\end{enumerate}

%% file: sections/related_work.tex
\section{Related Work}
\label{sec:related}

\paragraph{Long-Context RL Training.}
Recent work has demonstrated that reinforcement learning can improve long-context understanding beyond what supervised fine-tuning alone achieves. LoongRL~\citep{loongrl} composes short multi-hop QA pairs into long-context tasks through KeyChain, inducing reasoning patterns that transfer from 16K training to 128K evaluation. LongRLVR~\citep{longrlvr} identifies exponential gradient attenuation caused by sparse final-answer rewards and proposes dense context evidence rewards to mitigate it. QwenLong-L1.5~\citep{qwenlong} reports strong results through atomic-fact-based data synthesis, task-balanced RL with task-specific advantage estimation, and a memory-augmented architecture for ultra-long contexts. Despite their progress, these approaches focus on retrieval path injection, with task types largely restricted to QA variants and rewards often reduced to binary EM or accuracy signals. Capabilities such as summarization, ranking, aggregation, and structured reasoning therefore receive less explicit supervision. Our work addresses this gap with a capability-oriented dataset spanning 9 reward types, as described in Section~\ref{sec:data}.

\paragraph{GRPO Variants for Multitask RL.}
RLVR avoids the need to train reward models for long-context evaluation because rule-based reward functions such as EM, F1, and NDCG provide objective signals independent of the model being evaluated. GRPO~\citep{grpo} has become a common algorithm in this setting, estimating advantages as $A_i^u = (r_i - \mu_u)/\sigma_u$. This formulation, however, introduces biases that several variants seek to mitigate.

% \textit{Difficulty bias and its corrections.} Dr.\ GRPO~\citep{drgrpo} observes that dividing by the per-prompt standard deviation $\sigma_u$ can inflate advantages for both easy and hard prompts while suppressing medium-difficulty samples. It removes $\sigma_u$ to concentrate gradients on more informative examples. \red{The idea of reweighting training signals according to sample difficulty has a longer history in supervised learning. Focal Loss~\citep{focalloss} downweights well-classified examples so that the model concentrates on hard cases, and MiLe Loss~\citep{mileloss} extends this principle to language model pretraining by using prediction entropy to identify and emphasize difficult-to-learn tokens. Drawing on a similar intuition,} F-GRPO~\citep{fgrpo} applies Focal-style downweighting to easy prompts in the RL setting, but it estimates difficulty directly from raw rewards, which becomes less reliable under heterogeneous reward metrics. HA-DW~\citep{ha-dw} introduces history-aware adaptive weighting with finite-sample advantage bias analysis, although its historical mean is computed across tasks and may therefore conflate task difficulty with cross-task reward scale. These methods are helpful in single-task settings, but they do not fully address the interaction between difficulty bias and heterogeneous multitask rewards.

\textit{Difficulty bias and its corrections.} Dr.\ GRPO~\citep{drgrpo} observes that dividing by the per-prompt standard deviation $\sigma_u$ can inflate advantages for both easy and hard prompts while suppressing medium-difficulty samples. It removes $\sigma_u$ to concentrate gradients on more informative examples. F-GRPO~\citep{fgrpo} applies Focal-style downweighting to easy prompts~\citep{focalloss,mileloss}, but it estimates difficulty directly from raw rewards, which becomes less reliable under heterogeneous reward metrics. HA-DW~\citep{ha-dw} introduces history-aware adaptive weighting with finite-sample advantage bias analysis, although its historical mean is computed across tasks and may therefore conflate task difficulty with cross-task reward scale. These methods are helpful in single-task settings, but they do not fully address the interaction between difficulty bias and heterogeneous multitask rewards.

\textit{Cross-task scale normalization.} QwenLong-L1.5 adopts task-specific normalization, which normalizes advantages using the reward standard deviation within each task type and thereby reduces cross-task scale differences.

\textit{Orthogonal improvements.} DAPO~\citep{dapo} introduces asymmetric clipping for exploration, GPPO~\citep{gppo,su2026cegppocoordinatingentropygradientpreserving} propose gradient-preserving clipping, and ASPO~\citep{aspo} corrects importance sampling ratio mismatches for positive-advantage tokens. These methods primarily target policy optimization mechanics rather than the advantage estimation problem studied here.

\paragraph{Token-Level Importance Weighting for Long-Context Training.}\label{sec:token_reweighting}
A complementary line of work investigates how to weight individual tokens during long-sequence training. These studies observe that uniform-weight objectives can be suboptimal for long contexts and propose solutions including context-aware denoising objectives~\citep{cdt}, revised perplexity formulations for long-range utilization~\citep{longce}, and token-level loss weights that prioritize informative tokens~\citep{tokenweighting}. While these methods primarily operate in supervised settings, their insight that position- and content-dependent weighting can sharpen long-context learning signals is conceptually aligned with the difficulty-adaptive reweighting we apply at the response level in RL. Combining token-level weighting with RLVR remains a promising future direction.

%% file: sections/data.tex
\section{Data for Capability-Oriented Long-Context RLVR}
\label{sec:data}

Effective long-context data construction should start from the question of what capabilities long-context understanding requires, rather than from how to inject length into a prompt. Our data construction follows a capability-oriented framework, starting from the capabilities required by long-context models and designing tasks and rewards aligned with each. This section describes the design principles (Section~\ref{sec:data_principles}), dataset composition (Section~\ref{sec:data_composition}), data sources (Section~\ref{sec:data_sources}), construction pipeline (Section~\ref{sec:data_pipeline}), and data validity experiments (Section~\ref{sec:data_verify}).

\subsection{Design Principles}
\label{sec:data_principles}

Our data construction is guided by three principles.

\textbf{Capability orientation.} Rather than extending context length through additional retrieval hops or distractor density, we identify 9 core capabilities that long-context models should develop and define one task for each (Table~\ref{tab:dataset_composition}). The taxonomy is inspired by the capability dimensions proposed by LongBench Pro~\citep{longbenchpro}. Mapping data to these tasks helps the training process exercise a broad and balanced set of capabilities rather than repeatedly optimizing a single retrieval pattern.

\textbf{Reward alignment with task semantics.} Different long-context tasks carry different evaluation semantics, and collapsing them into a single reward metric can distort the training signal. Ranking tasks, for example, produce graded relevance judgments naturally measured by NDCG; forcing them into binary Exact Match would discard the partial-order structure of the output. We therefore assign each task its natural evaluation metric as the reward function, preserving more information in each reward signal and providing task-appropriate feedback during RL training.

% \textbf{Real document priority.} Template-based synthesis can generate data efficiently, but it may also introduce exploitable regularities. For example, when multiple short documents are concatenated and questions target only a subset of them, models may learn to exploit paragraph boundaries or formatting cues instead of developing genuine cross-document comprehension and summarization. We therefore prioritize real documents, including books, academic papers, legal filings, and financial reports, whenever feasible. Their information distribution, narrative structure, and linguistic variety provide a more natural source of difficulty. Synthetic data is used to supplement tasks where suitable real data is scarce, and even in those cases the synthesis is grounded in real source material to help preserve distributional authenticity.

\textbf{Real document priority.} Template-based synthesis can generate data efficiently but may introduce exploitable regularities. When multiple short documents are concatenated and questions target only a subset, models may learn to exploit paragraph boundaries or formatting cues rather than developing genuine cross-document comprehension. We therefore prioritize real documents, including books, academic papers, legal filings, and financial reports, whose information distribution, narrative structure, and linguistic variety provide a more natural source of difficulty. When annotated data in certain domains is scarce, we synthesize question-answer pairs from these real documents rather than generating the documents themselves, ensuring that the training signal remains grounded in authentic text.

\begin{figure}[t]
    \centering
    \begin{subfigure}[t]{0.48\textwidth}
        \centering
        \includegraphics[width=\linewidth]{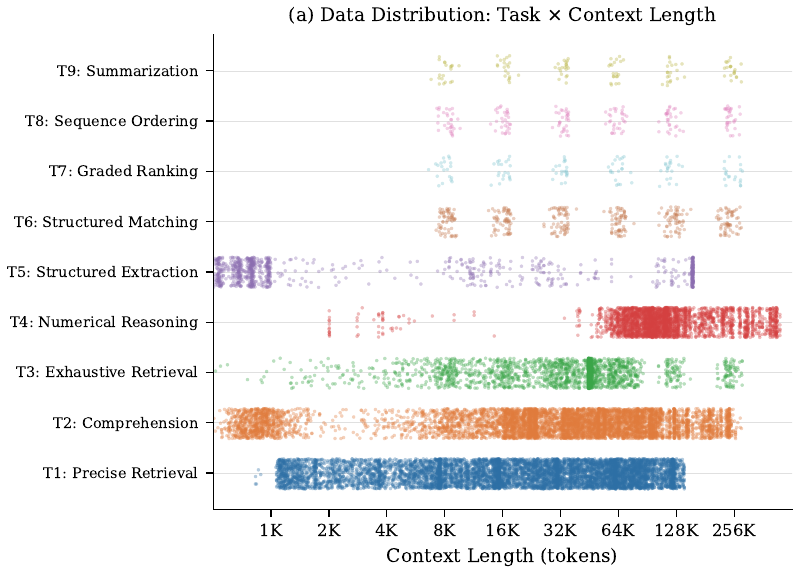}
        \caption{Joint distribution of tasks and context length. Each dot represents one training sample.}
        \label{fig:data_scatter}
    \end{subfigure}
    \hfill
    \begin{subfigure}[t]{0.48\textwidth}
        \centering
        \includegraphics[width=\linewidth]{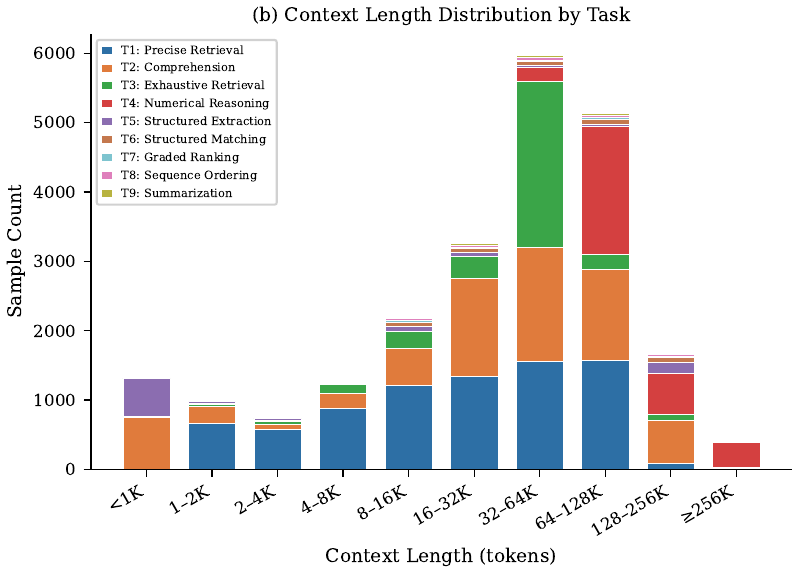}
        \caption{Context length histogram with log$_2$ scale bins, colored by task.}
        \label{fig:length_hist}
    \end{subfigure}
    \caption{Dataset distribution visualizations. The nine tasks T1 through T9 cover a broad range of context lengths rather than clustering within a narrow band. Most samples fall in the 4K to 256K range where long-context reasoning is most needed, while shorter and longer segments remain represented.}
    \label{fig:data_overview}
\end{figure}

\begin{table}[t!]
\centering
\caption{Composition of the capability-oriented long-context RLVR dataset. Each task is paired with its natural evaluation metric as the reward function.}
\label{tab:dataset_composition}
\footnotesize
\setlength{\tabcolsep}{2pt}
\renewcommand{\arraystretch}{1.08}
\begin{tabular}{clrrll}
\toprule
\textbf{Task} & \textbf{Reward Type} & \textbf{Samples} & \textbf{Ratio}
  & \textbf{Reward Function} & \textbf{Core Capability} \\
\midrule
T1 & EM
  & 7{,}908  & 34.4\% & Exact match
  & Precise long-range information retrieval \\
T2 & Accuracy
  & 6{,}808  & 29.6\% & Multiple-choice accuracy
  & Evidence-grounded comprehension and reasoning \\
T3 & F1
  & 3{,}478  & 15.1\% & Token F1
  & High-recall exhaustive retrieval and verification \\
T4 & \texttt{math\_verify}
  & 3{,}054  & 13.3\% & Math verification
  & Numerical extraction and quantitative reasoning \\
T5 & {IoU}
  & 937      & 4.1\%  & {IoU-based structured match}
  & Multi-table structured extraction \\
T6 & SubEM
  & 360      & 1.6\%  & Substring match
  & Fragment-level structured matching and induction \\
T7 & NDCG
  & 120      & 0.5\%  & Ranking quality
  & Dimension-quantified retrieval and graded ranking \\
T8 & Pairwise
  & 180      & 0.8\%  & Pairwise comparison
  & Sequence reconstruction and ordering \\
T9 & Summary
  & 120      & 0.5\%  & ROUGE-L
  & Long document summarization \\
\midrule
\textbf{} & \textbf{Total}
  & \textbf{22{,}965} & \textbf{100\%} & & \\
\bottomrule
\end{tabular}
\end{table}

\subsection{Overview of the Dataset}
\label{sec:data_composition}

The dataset contains \textbf{22,965} samples covering 9 tasks, with context lengths ranging from 0.1K to 256K tokens. Figure~\ref{fig:data_overview} visualizes the joint distribution of tasks and context lengths, showing broad coverage across both axes. Table~\ref{tab:dataset_composition} provides the full breakdown.

The task distribution is intentionally non-uniform. T1, T2, T3, and T4 constitute more than 90\% of the samples and form the training backbone because they cover fundamental long-context capabilities and benefit from abundant real-document sources. T6, T7, T8, and T9 together account for less than 4\% of the dataset because high-quality source material that naturally supports these formats is relatively scarce. We prioritize data quality over volume for these tasks, since retaining the natural reward form for each task contributes to complete capability coverage even at smaller scale.

\begin{figure}[t!]
    \centering
    \includegraphics[width=\linewidth]{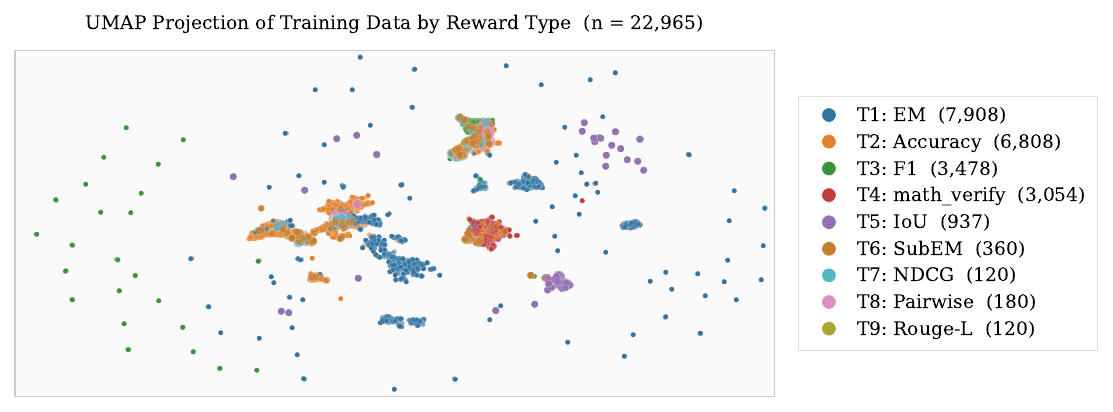}
    \caption{UMAP projection of 22{,}965 training samples. Different tasks occupy distinct regions in the embedding space, reflecting the semantic diversity of the capability-oriented dataset.}
    \label{fig:umap_dist}
\end{figure}

To further illustrate the semantic diversity of the dataset, Figure~\ref{fig:umap_dist} presents a UMAP projection of all training samples, colored by task and reward type. Samples from different tasks form well-separated clusters, confirming that the 9 reward types correspond to genuinely distinct semantic regions.

\begin{table}[t]
\centering
\caption{Mapping from data sources to tasks and reward types. Sample counts and capability descriptions are provided in Table~\ref{tab:dataset_composition}.}
\label{tab:data_sources}
\small
\setlength{\tabcolsep}{4pt}
\renewcommand{\arraystretch}{1.08}
\resizebox{\textwidth}{!}{%
\begin{tabular}{llll}
\toprule
\textbf{Origin} & \textbf{Source} & \textbf{Task} & \textbf{Reward} \\
\midrule
% \multirow{4}{*}{Synthetic}
\multirow{4}{*}{Synthetic}
  & \makecell[l]{Evidence integration QA from {Gutenberg~\citep{gutenberg}}, arXiv CC0~\citep{arxiv}, \\ \qquad \qquad and {PMC~\citep{pmc}}} & T2 & Accuracy \\
  & Rule induction reasoning with multiple choice questions & T2 & Accuracy \\
  & Dialogue memory and tracking from BEAM~\citep{beam} and Oolong~\citep{oolong} & T2 & Accuracy \\
  & Needle in a haystack samples & T1 & EM \\
\midrule
\multirow{10}{*}{Open source}
  & CLongEval~\citep{clongeval} from novels, news, tables, and related sources & T1 & EM \\
  & Filtered WikiHop~\citep{wikihop} & T1 & EM \\
  & {Gaokao Chinese~\citep{gaokao}}, LexSum~\citep{lexsum}, and LongBench Pro~\citep{longbenchpro} & T2 & Accuracy \\
  & LongBench Pro~\citep{longbenchpro} and CAIL2018~\citep{cail2018} & T3 & F1 \\
  & Financial QA~\citep{financialqa} & T4 & \texttt{math\_verify} \\
  & MultiTableQA~\citep{multitableqa} & T5 & IoU \\
  & LongBench Pro clustering, consistency, and rule induction subsets & T6 & SubEM \\
  & LongBench Pro ranking subset & T7 & NDCG \\
  & LongBench Pro ordering and frequency analysis subsets & T8 & Pairwise \\
  & LongBench Pro summarization subset & T9 & Summary \\
\bottomrule
\end{tabular}%
}
\end{table}

\subsection{Data Sources}
\label{sec:data_sources}

The dataset draws from two complementary pools, with roughly 14K curated open-source samples adapted from existing long-context corpora with available annotations and roughly 9K synthetic samples whose QA pairs are generated from real source documents. Table~\ref{tab:data_sources} maps each source to its task and reward type.

% \subsubsection{Synthetic Data}
% \label{sec:synth_data}
\subsubsection{Synthetic Data}
\label{sec:synth_data}

Among the 9 tasks, T2 is the primary target for large-scale synthesis because its multiple-choice format enables both reliable automatic generation and automatic verification, and its accuracy-based reward tends to reduce hallucinated labels during generation. Tasks with more complex reward formats, such as token-level F1, structured IoU, and ranking-based NDCG, are substantially harder to verify automatically during synthesis and are therefore covered through the open-source track (Section~\ref{sec:open_data}). In all cases, the source documents are real-world texts; what is generated is the question-answer pair, not the context itself. To capture diverse reasoning patterns within T2, we construct three categories of T2 samples.
 
\begin{itemize}
\item First, evidence integration QA, requires the model to locate and synthesize evidence from multiple scattered passages in long documents. Source texts are real documents drawn from Project Gutenberg books, arXiv CC0 academic papers, and PMC Open Access biomedical articles (Table~\ref{tab:data_sources}); a synthesis model generates questions and answers from these documents. All synthetic samples, including those from the other two T2 categories described below, undergo the multi-stage quality control detailed in Section~\ref{sec:data_pipeline}.

\item Second, rule induction reasoning, asks the model to infer an underlying rule from a small number of examples embedded in task-irrelevant natural language contexts and then apply the induced rule to a new case.
 
\item Third, dialogue memory and tracking, uses multi-turn dialogues with more than 50 turns and 30K tokens. Questions target information stated by a specific speaker in an earlier turn. 

\end{itemize}

We additionally synthesize approximately 4K needle-in-a-haystack samples for T1 with an exact match reward; like T2, the binary nature of T1's reward makes automatic verification straightforward. These T1 samples serve as a calibration anchor that helps the model retain basic long-range retrieval ability during training.

\subsubsection{Open-Source Data}
\label{sec:open_data}

% The remaining roughly 14K samples are curated from established corpora and mapped to the task and reward function that best matches their capability semantics. Table~\ref{tab:data_sources} lists the sources by task and reward type. Across the open-source subsets, the data covers domains such as legal case law, financial filings, literary fiction, and multi-turn dialogues, providing distributional coverage that complements the synthetic portion. All open-source samples undergo quality filtering, and samples with ambiguous labels or unanswerable questions are excluded.
The remaining roughly 14K samples are curated from established corpora and mapped to the task and reward function that best matches their capability semantics (Table~\ref{tab:data_sources}). The data covers domains such as legal case law, financial filings, literary fiction, and multi-turn dialogues, providing domain and format diversity that complements the synthetic portion. Unlike the synthetic track, this curated track starts from existing long-context samples with human-verified labels. {These samples undergo compatibility filtering and reward format standardization to ensure RLVR compatibility, as detailed in Section~\ref{sec:data_pipeline}.}

\subsection{Data Construction Pipeline}
\label{sec:data_pipeline}

\begin{figure}[t]
    \centering
    \includegraphics[width=\linewidth]{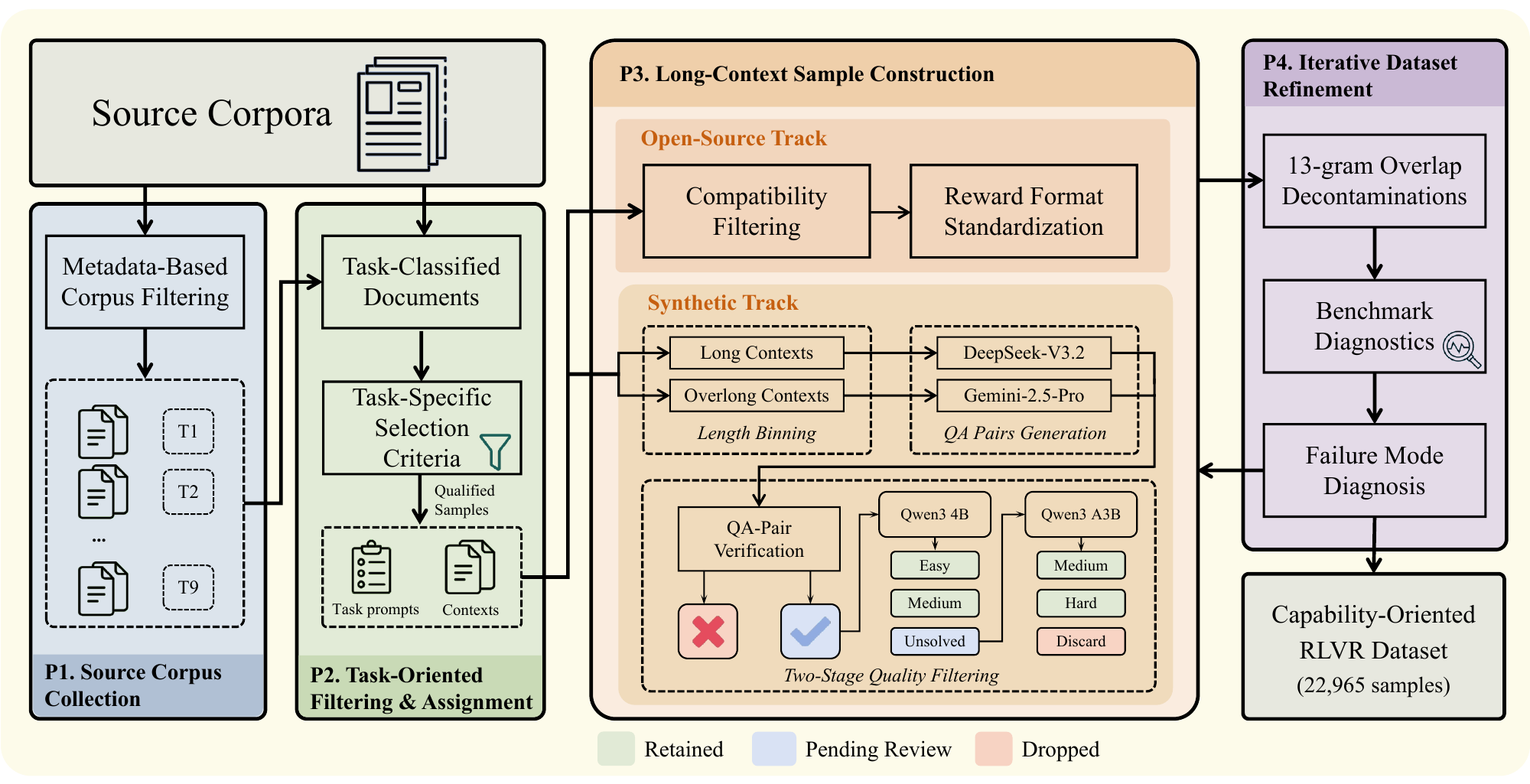}
    \caption{Overview of our four-phase Capability-Oriented RLVR Dataset construction pipeline.} 
    \label{fig:pipeline}
\end{figure}

{The entire dataset, spanning both the open-source and synthetic tracks, is produced through a unified four-phase pipeline illustrated in Figure~\ref{fig:pipeline}.}

\textbf{P1 Source corpus collection.} Guided by the 9-task capability taxonomy (Section~\ref{sec:data_principles}), this phase collects two categories of source corpora through careful manual curation. The first consists of annotated open-source datasets such as CLongEval, WikiHop, LongBench Pro, and FinancialQA (Table~\ref{tab:data_sources}), which provide long-context QA annotations for the open-source track. The second consists of unannotated real-world documents for the synthetic track, including Project Gutenberg books, arXiv CC0 papers, PMC Open Access articles, and multi-turn dialogues from BEAM and Oolong. Corpora are selected to maximize diversity in domain, document structure, and length distribution across the target tasks.

{\textbf{P2 Task-oriented filtering and assignment.} This phase determines the task assignment for every sample or document by applying task-specific selection criteria. For open-source datasets, we select subsets whose evaluation semantics align with a particular task. For example, CLongEval samples that require pinpointing a single factual span are assigned to T1, while CAIL2018 samples that require aggregating multiple legal provisions are assigned to T3. For unannotated documents, filtering operates at the document level based on structural properties required by each task. The dialogue memory subcategory of T2, for instance, retains only multi-turn conversations containing speaker-specific, traceable statements and excludes dialogues with fewer than 50 turns or total length below 30K tokens. At the end of this phase, every retained sample or document has been assigned to exactly one task.}

{\textbf{P3 Sample construction.} This phase converts filtered materials from P2 into training-ready RLVR samples through two parallel tracks.}

{\emph{Open-source track.} Samples from established corpora already carry human-verified annotations, but these are not always directly compatible with the RLVR reward format. Construction proceeds in two steps. Compatibility filtering first examines whether each sample's annotation can be transformed into a reward-computable answer for its assigned task, discarding incompatible samples. For T1, which uses exact match, we retain only samples whose reference answers are short spans, because longer answers introduce matching ambiguity. Reward format standardization then converts the remaining annotations into the format expected by each task's reward function. T4 numerical reasoning samples, for example, have their answers reformulated as expressions parsable by \texttt{math\_verify}, and T7 ranking samples have their annotations restructured as ordered lists evaluable by NDCG.}

\emph{Synthetic track.} Filtered real documents are converted into question-answer pairs through three steps. First, documents are tokenized and assigned to length bins so that the appropriate synthesis model can be selected for each bin. We use DeepSeek V3.2~\citep{deepseekv32} for sequences within its context window and Gemini-2.5-Pro~\citep{gemini25} for longer samples. Second, the synthesis model generates candidate questions and answers from a task-specific prompt, following a three-step procedure in which the model identifies the source language, constructs questions with task-specific constraints including plausible distractors and at least four options, and performs a self-check for task compliance, answer accuracy, and absence of ambiguity (Appendix~\ref{app:shared_prompt}). Third, each generated sample undergoes two stages of quality filtering. In the first stage, QA-pair verification by Gemini-2.5-Pro reviews the full source document to confirm that the question is well-defined, the designated answer is unambiguously supported, distractors are plausible but clearly incorrect, and no hallucinated content is present (Appendix~\ref{app:verification_prompt}); samples that fail any check are discarded. In the second stage, multi-stage verification validates answer-label correctness by evaluating each sample with models of different capability levels. Because the synthetic data covers only T1 and T2, both of which use binary rewards (EM and multiple-choice Accuracy yield $r_i \in \{0, 1\}$), the pass rate is simply the fraction of rollouts that answer correctly. In Stage~1, each sample is evaluated by Qwen3-4B-Thinking-2507 with $G{=}8$ rollouts. Samples with pass rate above $0.75$ are labeled \emph{easy}, those between $0.5$ and $0.75$ are labeled \emph{medium}, and those below $0.5$ are labeled \emph{unsolved}. Unsolved samples are forwarded to Stage~2, where Qwen3-30B-A3B with $G{=}8$ rollouts reclassifies them as \emph{medium} (pass rate above $0.75$), \emph{hard} (pass rate between $0.25$ and $0.75$), or \emph{quality-insufficient} (pass rate below $0.25$). Samples labeled quality-insufficient are discarded, as their consistently low pass rates under both model scales suggest label noise rather than genuine difficulty. Across the full training set evaluated by the 4B model, approximately 31\% of samples achieve a perfect pass rate, 58\% fall in the intermediate range, and 11\% are never solved, yielding an approximate easy-to-medium-to-hard ratio of $3{:}6{:}1$.

{\textbf{P4 Iterative refinement.} After P3, samples from both tracks are merged into a unified dataset. We refine the dataset through multiple version iterations. In each iteration, we first apply 13-gram overlap filtering between training queries and all benchmark queries, discarding any training sample that shares a 13-gram sub-string with an evaluation query to prevent data contamination. We then train a model on the filtered dataset and evaluate it on downstream benchmarks to diagnose weak capability dimensions. When a task exhibits stagnant improvement despite sufficient sample volume, we inspect the corresponding training samples for systematic issues such as reward hacking, answer ambiguity, or exploitable formatting shortcuts, and remove the offending samples. When a capability dimension shows limited gains due to insufficient training signal, we repeat P1 through P3 to supplement the dataset with additional samples from newly identified corpora or freshly synthesized QA pairs. This cycle is repeated until benchmark performance and data quality stabilize. The complete set of generation prompts is provided in Appendix~\ref{app:prompts}, and the verification prompt is provided in Appendix~\ref{app:verification_prompt}.}

\begin{table}[t]
\centering
\begin{threeparttable}
\caption{Data effectiveness validation across model scales. Bold marks the best result within each scale and underlining marks the second best. In ``Flagship Reasoning Models'' and ``Lightweight Reasoning Models'', bold indicates the best performance within each respective group.}
\label{tab:data_effectiveness}
\footnotesize
\setlength{\tabcolsep}{2.5pt}
\renewcommand{\arraystretch}{1.00}
\begin{tabular}{clccccccc}
\toprule
\textbf{Scale} & \textbf{Model} & \textbf{Avg.} & \textbf{DocMath}
  & \textbf{LBV2} & \textbf{Frames} & \textbf{MRCR}
  & \textbf{CorpusQA} & \textbf{LBV1-QA} \\
\midrule
\multicolumn{9}{c}{\textit{Flagship Reasoning Models}} \\
\midrule
  & GPT-5 \tnote{$\dagger$}
    & \textbf{74.74} & \textbf{67.62} & 62.82 & \textbf{84.59} & 77.29 & \textbf{81.56} & \textbf{73.70} \\
  & Gemini-2.5-Pro \tnote{$\dagger$}
    & 72.40 & 62.38 & \textbf{65.72} & 74.51 & \textbf{79.92} & 80.62 & 71.28 \\
  & Qwen3-Max-Thinking-Preview \tnote{$\dagger$}
    & 69.43 & 64.12 & 57.89 & 77.93 & 71.24 & 74.69 & 70.71 \\
  & DeepSeek-R1-0528 \tnote{$\dagger$}
    & 68.67 & 63.44 & 59.48 & 76.86 & 64.88 & 77.50 & 69.90 \\
  & Qwen3-235B-A22B-Thinking-2507 \tnote{$\dagger$}
    & 68.45 & 65.75 & 57.46 & 75.12 & 66.17 & 75.31 & 70.90 \\
\midrule
\multicolumn{9}{c}{\textit{Lightweight Reasoning Models}} \\
\midrule
  & Gemini-2.5-Flash-Thinking \tnote{$\dagger$}
    & 68.73 & 64.75 & 56.77 & 65.78 & 78.84 & \textbf{79.38} & 66.86 \\
  & GPT-OSS-120B \tnote{$\dagger$}
    & 58.55 & 61.25 & 47.01 & 72.69 & 39.68 & 64.38 & 66.30 \\
  & GPT-5-Nano \tnote{$\dagger$}
    & 57.06 & 63.88 & 43.74 & 73.54 & 43.88 & 50.31 & 67.10 \\
% \midrule
% \multicolumn{9}{c}{\textit{Qwen3.5 Series Models}} \\
% \midrule
  & {Qwen3.5-4B} 
    & 68.81 & 65.87 & 45.92 & 70.87 & 96.09 & 70.52 & 63.60 \\
  & {Qwen3.5-35B-A3B} 
    & \textbf{74.91} & \textbf{68.87} & \textbf{59.44} & \textbf{74.88} & \textbf{97.37} & 78.72 & \textbf{70.20} \\
\midrule
\multicolumn{9}{c}{\textit{Qwen3 Series Models}} \\
\midrule

\multirow{4}{*}{4B}
  & QwenLong-L1.5 \textit{(w. GRPO)} \tnote{$\dagger$}
    & 56.1 & \second{61.3} & 44.3 & \second{67.1} & 40.9 & 58.8 & 64.1 \\
  & QwenLong-L1.5 \textit{(w. AEPO)} \tnote{$\dagger$}
    & \second{59.4} & \textbf{62.5} & \textbf{47.9} & \textbf{67.4} & \second{47.9} & \second{64.7} & \second{65.8} \\
  \cmidrule(l){2-9}
  & Qwen3-4B-Thinking-2507
    & 53.0 & 61.0 & 40.2 & 64.4 & 38.4 & 49.9 & 64.0 \\
  \rowcolor{RowHighlight}
  % & \quad + GRPO \textit{(ours)}\tnote{$\ddagger$}
  & \textbf{GoLongRL-4B} \textit{(w. GRPO)}\tnote{$\ddagger$}
    & \textbf{62.2} & \textbf{62.5} & \second{45.5} & 66.6
    & \textbf{67.5} & \textbf{65.1} & \textbf{65.9} \\
\midrule
\multirow{4}{*}{30B}
  & QwenLong-L1.5 \textit{(w. GRPO)} \tnote{$\dagger$}
    & 67.2 & 65.1 & \textbf{55.3} & \second{71.4} & 66.9 & \second{76.9} & \second{67.9} \\
  & QwenLong-L1.5 \textit{(w. AEPO)}
    & \textbf{71.2} & \textbf{66.4} & \second{55.2} & \textbf{74.5} & \textbf{82.5} & \textbf{80.9} & 67.7 \\
  \cmidrule(l){2-9}
  & Qwen3-30B-A3B-Thinking-2507
    & 60.1 & 63.3 & 48.7 & 70.2 & 41.6 & 70.5 & 66.5 \\
  \rowcolor{RowHighlight}
  % & \quad + GRPO \textit{(ours)}
  & \textbf{GoLongRL-30B-A3B} \textit{(w. GRPO)}
    & \second{69.8} & \second{65.3} & 55.1 & \textbf{74.5}
    & \second{81.6} & 73.6 & \textbf{68.7} \\
\bottomrule
\end{tabular}
\begin{tablenotes}
  \footnotesize
  \item[$\dagger$] {Flagship, lightweight, and QwenLong results reported from \citet{qwenlong}; Qwen3.5 and all other results evaluated under our unified protocol.}
  \item[$\ddagger$] Trained on a randomly sampled 8K subset of the full dataset. See Section~\ref{sec:data_verify} for details.
\end{tablenotes}
\end{threeparttable}
\end{table}

\subsection{Data Validity Verification}
\label{sec:data_verify}

To evaluate the independent effect of the capability-oriented data scheme, we conduct validation experiments at two model scales while fixing the algorithm to vanilla GRPO, thereby decoupling the data contribution from algorithmic changes. The model configurations, hyperparameters, and evaluation protocols follow Section~\ref{sec:experiments}. Because long-context RLVR is computationally expensive, the 4B model is trained on a randomly sampled 8,000-example subset of the full dataset, enabling efficient iteration and ablation. For consistency, all later 4B-scale experiments use the same subset. The 30B model is trained on the full dataset to evaluate scalability. As the 30B MoE architecture introduces additional optimization variables unrelated to advantage estimation, the algorithmic ablation is conducted separately in the more controlled 4B dense setting (Section~\ref{sec:experiments}).

% Table~\ref{tab:data_effectiveness} provides evidence for the independent contribution of the capability-oriented dataset. Both comparison groups use GRPO, so performance differences can be attributed primarily to data construction. At the 4B scale, GRPO trained on our dataset reaches an average of 62.2, exceeding GRPO on QwenLong-L1.5 data by 6.1 points. At the 30B scale, the advantage remains positive (69.8 vs. 67.2). Notably, vanilla GRPO trained on our data already approaches QwenLong-L1.5 with its specialized AEPO algorithm (62.2 vs. 59.4 at 4B and 69.8 vs. 71.2 at 30B), suggesting that data construction and algorithm design are complementary drivers of long-context RL performance. {\color{blue} Table~\ref{tab:data_effectiveness} further includes the newer-generation Qwen3.5 series for cross-generational comparison. GoLongRL-30B-A3B, built on the older Qwen3 backbone, already surpasses Qwen3.5-4B (69.8 vs. 68.81) on four of six benchmarks, with the remaining gap concentrated on MRCR where Qwen3.5 benefits from stronger base retrieval capability.} These results suggest that capability-oriented multitask coverage and heterogeneous reward design provide richer training signals than approaches that focus primarily on retrieval-based tasks with a single reward format. {\color{blue} As a post-training recipe orthogonal to base model improvements, GoLongRL can be applied to stronger backbones, suggesting room for further gains.}

Table~\ref{tab:data_effectiveness} provides evidence for the independent contribution of the capability-oriented dataset. Both comparison groups use GRPO, so performance differences can be attributed primarily to data construction. At the 4B scale, GRPO trained on our dataset reaches an average of 62.2, exceeding GRPO on QwenLong-L1.5 data by 6.1 points. At the 30B scale, the advantage remains positive (69.8 vs. 67.2). Notably, vanilla GRPO trained on our data already approaches QwenLong-L1.5 with its specialized AEPO algorithm (62.2 vs. 59.4 at 4B and 69.8 vs. 71.2 at 30B), suggesting that data construction and algorithm design are complementary drivers of long-context RL performance. These results suggest that capability-oriented multitask coverage and heterogeneous reward design provide richer training signals than approaches that focus primarily on retrieval-based tasks with a single reward format. As a post-training recipe orthogonal to base model improvements, GoLongRL can be applied to stronger backbones, suggesting room for further gains.

\begin{table}[t]
\centering
\begin{threeparttable}
\caption{Dataset iteration. Each version is trained on an 8K randomly sampled subset from its data pool with vanilla GRPO at the 4B scale.}
\label{tab:iteration_history}
\small
\setlength{\tabcolsep}{4pt}
\renewcommand{\arraystretch}{1.1}
\begin{tabular}{lrcccccccc}
\toprule
\textbf{Version} & \textbf{Size} & \textbf{Avg.} & \textbf{DocMath} & \textbf{LBV2} & \textbf{Frames} & \textbf{MRCR} & \textbf{CorpusQA} & \textbf{LBV1-QA} \\
\midrule
V1 & 9{,}630 & 52.8 & 59.8 & 42.5 & 62.1 & 38.8 & 50.2 & 63.6 \\
V2 & 17{,}729 & 55.0 & 60.0 & 43.7 & 64.4 & 40.7 & 57.8 & 63.5 \\
V3\tnote{$*$} & 22{,}965 & \textbf{62.2} & \textbf{62.5} & \textbf{45.5} & \textbf{66.6} & \textbf{67.5} & \textbf{65.1} & \textbf{65.9} \\
\bottomrule
\end{tabular}
\begin{tablenotes}
\footnotesize
\item[$*$] V3 corresponds to the final dataset reported in the GoLongRL-4B row of Table~\ref{tab:data_effectiveness}.
\end{tablenotes}
\end{threeparttable}
\end{table}

Table~\ref{tab:iteration_history} records the training results at three key dataset versions during the P4 iterative refinement process (Section~\ref{sec:data_pipeline}). After training on the initial version V1 (9.6K pool), single-document comprehension and basic retrieval metrics were already close to their final levels, but cross-document reasoning (CorpusQA 50.2) and long-range context memory (MRCR 38.8) lagged noticeably behind, pointing to uneven task coverage. We returned to the P1 through P3 pipeline to broaden task-type coverage, producing V2 (17.7K pool). CorpusQA improved to 57.8 (+7.6), yet MRCR stalled at 40.7, suggesting that this capability dimension responds weakly to general data expansion. Following this diagnosis, we supplemented V2 with targeted multi-hop reasoning and context-memory samples, yielding the final version V3 (23.0K pool). MRCR jumped from 40.7 to 67.5 (+26.8) and all remaining metrics improved, bringing the average to 62.2. Notably, V2 to V3 added only 5.2K samples yet produced an average gain of +7.2, compared with +2.2 from the 8.1K-sample expansion of V1 to V2, indicating that diagnosis-driven supplementation achieves higher data efficiency than undifferentiated expansion. This precise diagnosis and targeted repair is made possible by the capability-oriented task taxonomy (Section~\ref{sec:data_principles}), which provides actionable evaluation dimensions for each iteration.

%% file: sections/algorithm.tex
\section{TMN-Reweight for Multitask Long-Context RL}
\label{sec:algorithm}

The capability-oriented dataset described in Section~\ref{sec:data} employs 9 reward functions with different numerical scales and variance profiles. Training on this mixture with standard GRPO can give rise to two optimization issues. This section first reviews GRPO and its variants in Section~\ref{sec:grpo_preliminary}, then analyzes the two issues in Section~\ref{sec:grpo_defects}, introduces TMN-Reweight in Section~\ref{sec:tmn_reweight}, and compares it with existing approaches in Section~\ref{sec:algo_compare}.

\subsection{Preliminaries for GRPO and Its Variants}
\label{sec:grpo_preliminary}

We use Group Relative Policy Optimization (GRPO)~\citep{grpo} as the base RL algorithm. Unlike PPO~\citep{ppo}, which requires a separate value network, GRPO estimates advantages by normalizing rewards within a sampled group of responses, eliminating the value network and reducing the cost of long-context training.

\textbf{GRPO objective.} For each prompt $u$, which contains a context $c$ and a question $q$, GRPO samples a group of $G$ candidate responses $\{o_i\}_{i=1}^{G}$ from the current policy $\pi_{\theta_{\text{old}}}$, collects their rewards $\{r_i\}_{i=1}^{G}$, and computes the response advantage through group-level z-score normalization.
\begin{equation}
\label{eq:grpo_adv}
    A_i^{u} = \frac{r_i - \mu_u}{\sigma_u + \delta}, \quad \mu_u = \frac{1}{G}\sum_{j=1}^{G} r_j, \quad \sigma_u = \sqrt{\frac{1}{G-1}\sum_{j=1}^{G}(r_j - \mu_u)^2}
\end{equation}
Here, $\mu_u$ and $\sigma_u$ denote the mean and standard deviation of rewards within the group, and $\delta$ is a small constant for numerical stability. The policy is then updated by maximizing the clipped surrogate objective.
\begin{equation}
\label{eq:grpo_obj}
    \mathcal{J}_{\text{GRPO}}(\theta) = \mathbb{E}_{u \sim \mathcal{D},\, \{o_i\}_{i=1}^{G} \sim \pi_{\theta_{\text{old}}}(\cdot \mid u)} \left[ \frac{1}{G}\sum_{i=1}^{G} \frac{1}{|o_i|}\sum_{t=1}^{|o_i|} \min\!\Big(\rho_{i,t}(\theta)\, A_i^{u},\;\; \text{clip}\big(\rho_{i,t}(\theta),\, 1{-}\varepsilon,\, 1{+}\varepsilon\big)\, A_i^{u}\Big) \right]
\end{equation}
The importance sampling ratio is $\rho_{i,t}(\theta) = \pi_{\theta}(o_{i,t} \mid u, o_{i,<t}) \,/\, \pi_{\theta_{\text{old}}}(o_{i,t} \mid u, o_{i,<t})$ for token $t$ in response $i$. The clipping range $\varepsilon$ stabilizes the update by limiting large policy shifts. Following recent practice, we omit the KL divergence penalty term $\beta\,\mathbb{D}_{\text{KL}}(\pi_\theta \| \pi_{\text{ref}})$ because the model distribution is expected to move away from the initial policy during long-context RL.

\textbf{Dr.\ GRPO.} Dr.\ GRPO~\citep{drgrpo} identifies two biases in standard GRPO. First, division by the per-prompt standard deviation $\sigma_u$ in Eq.~\ref{eq:grpo_adv} creates question-level difficulty bias. Prompts where all rollouts receive similar rewards, whether mostly correct or mostly incorrect, can have small $\sigma_u$ and therefore inflated normalized advantages. Second, averaging the sample loss by response length $|o_i|$ in Eq.~\ref{eq:grpo_obj} creates response length bias and can underweight longer responses. Dr.\ GRPO mitigates these issues by removing $\sigma_u$ from advantage computation and replacing per-sample length normalization with a global constant $L$, typically the maximum completion length.
\begin{equation}
\label{eq:drgrpo}
    \mathcal{J}_{\text{Dr.GRPO}}(\theta) = \mathbb{E}\left[\frac{1}{G}\sum_{i=1}^{G}\frac{1}{L}\sum_{t=1}^{|o_i|} \min\!\Big(\rho_{i,t}\, A_i^{u},\;\; \text{clip}\big(\rho_{i,t},\, 1{-}\varepsilon,\, 1{+}\varepsilon\big)\, A_i^{u}\Big)\right]
\end{equation}
where $A_i^{u} = r_i - \mu_u$. This formulation concentrates more gradient weight on medium-difficulty prompts and provides more consistent per-token learning signals across response lengths.

\textbf{Multitask training objective.} In our setting, the training data spans $K$ task types $\{\mathcal{T}_k\}_{k=1}^{K}$, each associated with its own reward function $r^{(k)}$, including EM, Accuracy, F1, and NDCG. We sample prompts uniformly from the full dataset, so each task's effective 
weight is proportional to its share of the training data 
(Table~\ref{tab:dataset_composition}). The overall training objective is
\begin{equation}
\label{eq:multi_task_obj}
    \mathcal{J}(\theta) = \mathbb{E}_{k \sim \mathcal{P}(\mathcal{T})} \;\mathbb{E}_{u \sim \mathcal{T}_k,\, \{o_i\} \sim \pi_{\theta_{\text{old}}}(\cdot \mid u)} \left[{\frac{1}{\sum_{j=1}^{G}|o_j|}}\sum_{i=1}^{G} \sum_{t=1}^{|o_i|} \mathcal{L}_{\text{clip}}\!\big(\rho_{i,t},\, \tilde{A}_i^{(k)}\big)\right]
\end{equation}
where $\mathcal{P}(\mathcal{T})$ is the task sampling distribution proportional to each task's sample proportion, $\tilde{A}_i^{(k)}$ is the modified advantage for response $i$ under task type $k$, and $\mathcal{L}_{\text{clip}}$ is the standard clipped surrogate loss. We normalize the per-token loss by the total number of tokens across all responses in the group rather than by individual response length, providing more uniform per-token learning signals across varying completion lengths. The central challenge is to compute $\tilde{A}_i^{(k)}$ so that heterogeneous rewards produce comparable gradient magnitudes across tasks while preserving useful within-task difficulty information.

\subsection{Problem Analysis}
\label{sec:grpo_defects}

Although GRPO and Dr.\ GRPO are effective in single-task RL settings, their advantage estimation strategies can interact poorly with heterogeneous multitask long-context training.

\textbf{Defect 1. Difficulty-induced advantage bias.} Dividing by the per-prompt standard deviation $\sigma_u$ amplifies advantages for both hard and easy prompts, while compressing advantages for medium-difficulty prompts whose rollouts exhibit greater outcome diversity. Medium-difficulty prompts are often the most informative because the model can sometimes solve them but does not yet do so reliably. Dr.\ GRPO addresses this issue by removing $\sigma_u$, which reduces the advantage to the raw deviation $r_i - \mu_u$. While this is a reasonable correction in single-task settings, in heterogeneous multitask training it exposes a second issue.

\textbf{Defect 2. Cross-task reward scale inconsistency.} Different tasks in our dataset use different evaluation metrics, including EM, F1, NDCG, and ROUGE-L. These metrics produce reward distributions with different scales and variance profiles. Once $\sigma_u$ is removed, tasks with higher reward variance can generate disproportionately large gradients and dominate optimization, while tasks with lower reward variance may receive weaker learning signals. In our dataset, for instance, F1-based retrieval tasks have reward distributions that differ substantially from multiple-choice Accuracy (binary-valued) tasks, and mixing them without normalization can distort the optimization trajectory.

These two defects call for different interventions. Defect~1 motivates removing per-prompt normalization to avoid over-amplifying prompts at extreme difficulty levels. Defect~2 motivates some form of normalization to align cross-task reward scales. Existing methods typically target only one side of this tension, and addressing one defect in isolation can exacerbate the other.

\subsection{TMN-Reweight}
\label{sec:tmn_reweight}

TMN-Reweight decouples scale normalization from difficulty correction through two transformations that define $\tilde{A}_i^{(k)}$ in Eq.~\ref{eq:multi_task_obj}.

\begin{figure}[t]
    \centering
    \includegraphics[width=\linewidth]{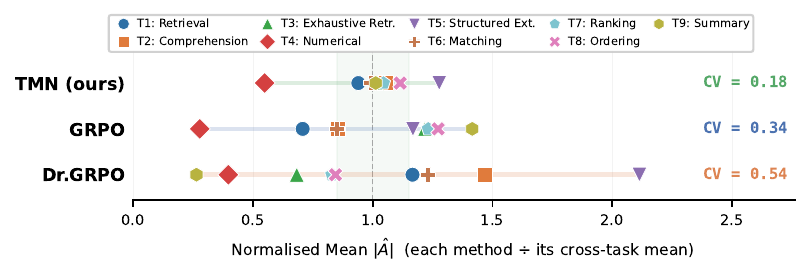}
    \caption{Normalised mean absolute advantage per task under three advantage estimation methods. For each method, per-task values are divided by the method-specific cross-task mean, so ideal uniformity corresponds to values near $1.0$. The shaded band marks the ${\pm}15\%$ convergence region. The coefficient of variation quantifies cross-task disparity, where lower values indicate more uniform gradient magnitudes.}
\label{fig:advantage_norms}
\end{figure}

\textbf{Step 1. Task-level Mean Normalization.} Instead of normalizing by the per-prompt standard deviation $\sigma_u$, which conflates scale correction with difficulty amplification, we normalize by a task-level aggregated standard deviation. Concretely, we first compute the within-group standard deviation $\sigma_u$ independently for each prompt $u$, then aggregate these per-prompt values across all prompts in the same task via root mean square
\begin{equation}
    \hat{A}_i^{u} = \frac{r_i - \mu_u}{\sigma_{\text{task}(i)} + \delta}, \quad \text{where } \sigma_{\text{task}(i)} = \sqrt{\frac{1}{|U_{\text{task}}|}\sum_{u \in U_{\text{task}}} \sigma_u^2}
\end{equation}
Here, $\sigma_{\text{task}(i)}$ is the root mean square of per-prompt standard deviations within the task of prompt $i$. This aggregation is theoretically motivated by a gradient norm analysis (Appendix~\ref{app:tmn_theory}), which shows that the per-task gradient magnitude scales with $\sqrt{\mathbb{E}_{u \sim \mathcal{T}_k}[\sigma_u^2]}$, of which $\sigma_{\text{task}(i)}$ is the empirical estimate. Because the denominator is shared across prompts in the same task, this operation reduces cross-task scale differences while preserving relative advantage magnitudes within each task. In contrast to Dr.\ GRPO, which removes normalization entirely, TMN replaces per-prompt normalization with per-task normalization and leaves within-task difficulty structure available for the second step.

Figure~\ref{fig:advantage_norms} provides quantitative support. We compute the mean absolute advantage per task under GRPO, Dr.\ GRPO, and TMN using a single inference pass, then normalize each method by its cross-task mean to isolate scale uniformity. Dr.\ GRPO yields a cross-task coefficient of variation of $0.54$, as removing the denominator allows high-variance tasks to dominate. GRPO reduces this to $0.34$ through per-prompt normalization, but cross-task disparity persists because $\sigma_u$ corrects within-prompt rather than between-task scale. TMN further reduces the coefficient of variation to $0.18$, placing all nine tasks within a narrow band around unity.

\textbf{Step 2. Difficulty-adaptive reweighting.} After TMN reduces cross-task scale discrepancies, the remaining issue is difficulty bias within each task. We estimate prompt difficulty with a smoothed pass rate and apply exponential weighting, following the intuition from HA-DW~\citep{ha-dw} that finite-sample advantage estimates can underestimate hard prompts.

A central design choice is how to estimate prompt difficulty. The naive estimate $\hat{p}_u = \frac{1}{G}\sum_i \mathbf{1}[r_i > 0]$ from $G$ rollouts exhibits high variance. With $G{=}8$, a hard prompt may have pass rate $0$ in one batch and $0.125$ in another. A batch-level pass rate across all prompts, on the other hand, discards prompt-level granularity. We address this by smoothing the per-prompt mean reward with a task-level baseline.
\begin{equation}
    \tilde{\mu}_u = \alpha \cdot \mu_u + (1 - \alpha) \cdot \mu_{\text{task}}, \quad \hat{p} = \frac{\sum_{i=1}^{G} \mathbf{1}[r_i > \tilde{\mu}_u]}{G}
\end{equation}
Here, $\mu_{\text{task}}$ is the average reward over prompts from the same task type within the current batch. The interpolation coefficient $\alpha$ controls the balance between prompt-level sensitivity and task-level stability. Values near $1$ preserve prompt-level granularity, while values near $0$ borrow more strength from the task-level estimate. The smoothing is performed strictly within task boundaries to avoid cross-task contamination.

The difficulty weight is computed as
\begin{equation}
    w = \exp(0.5 - \hat{p})
\end{equation}
When $\hat{p} < 0.5$, the prompt is relatively hard and $w > 1$, which amplifies rare positive samples representing successful exploration on difficult problems. When $\hat{p} > 0.5$, the prompt is relatively easy and $w < 1$, which reduces the optimization signal for problems the model already solves reliably.

\textbf{Four-quadrant gradient reallocation.}
Applying $w$ symmetrically to all trajectories introduces two concerns. For hard prompts, scaling up the dominant negative samples can create large opposing gradients and destabilize optimization; DeepSeek-V3.2 reports a related failure mode in its Off-Policy Sequence Masking analysis where divergent negative sequences mislead the update. For easy prompts, repeatedly reinforcing correct responses can concentrate probability mass and contribute to entropy collapse.

We address both concerns by applying $w$ asymmetrically according to the sign of the TMN advantage $\hat{A}_i^{u}$. A sample is considered positive when $\hat{A}_i^{u} > 0$, meaning its raw reward exceeds the group mean $\mu_u$, rather than the smoothed threshold $\tilde{\mu}_u$ used for pass rate estimation. These two thresholds serve distinct roles. The advantage sign determines gradient direction and should preserve standard group-relative semantics, reinforcing responses that exceed the policy's current expected performance. Replacing $\mu_u$ with $\tilde{\mu}_u$ would weaken within-group contrast, particularly on easy prompts where the task-level component could shift the threshold below $\mu_u$ and cause nearly all trajectories to receive positive advantage. The smoothed threshold $\tilde{\mu}_u$ provides a stabilized difficulty estimate that reduces variance from small group sizes, but this statistical benefit should not alter gradient direction.
\begin{equation}
    \tilde{A}_i = \begin{cases}
    \hat{A}_i^{u} \cdot w
      & \text{if } \hat{A}_i^{u} > 0 \\[4pt]
    \hat{A}_i^{u} \cdot \dfrac{1}{w}
      & \text{otherwise}
    \end{cases}
\end{equation}
This rule creates four gradient regimes. For hard prompts ($w > 1$), positive advantages are amplified to reinforce rare successful rollouts, while negative advantages are scaled down by $1/w$ to reduce destabilizing gradients, acting as a softer alternative to masking. For easy prompts ($w < 1$), positive advantages are attenuated to slow probability concentration on already-solved outputs, while negative advantages are amplified by $1/w$ to increase learning from unexpected failures.

ASPO~\citep{aspo} also treats positive and negative advantages asymmetrically, but through a different mechanism. ASPO modifies token-level importance sampling ratios for positive-advantage tokens to correct weight mismatch inside the clipping objective. TMN-Reweight does not modify the importance sampling ratio. Instead, it applies a prompt-level difficulty weight asymmetrically by scaling positive and negative advantages by $w$ and $1/w$, respectively.

Combining the two steps, TMN-Reweight is designed to mitigate both issues identified in Section~\ref{sec:grpo_defects}. Task-level normalization reduces cross-task reward scale discrepancies, while difficulty-adaptive reweighting adjusts the learning signal according to prompt difficulty.

\subsection{Comparison with Existing Approaches}
\label{sec:algo_compare}

We compare TMN-Reweight with existing GRPO variants along the two axes identified in Section~\ref{sec:grpo_defects}. GRPO does not directly address either issue. Dr.\ GRPO removes $\sigma_u$ to reduce difficulty bias but can allow high-variance tasks to dominate in multitask settings. F-GRPO applies Focal-style downweighting to easy prompts but estimates difficulty from raw rewards, which is less reliable across heterogeneous metrics. HA-DW introduces history-aware adaptive weighting but computes its historical mean across tasks, potentially conflating difficulty with cross-task reward scale. QwenLong-L1.5 normalizes advantages by task-level reward standard deviation, addressing cross-task scale but without explicit difficulty correction. Each method targets at most one of the two issues, and the correction may introduce sensitivity along the other axis.

TMN-Reweight combines task-level normalization for cross-task scale alignment with difficulty-adaptive four-quadrant reweighting for difficulty bias correction. The empirical results in Section~\ref{sec:experiments} suggest that these two components are complementary.

%% file: sections/experiments.tex
\section{Experiments}
\label{sec:experiments}
\subsection{Setup}
% TODO icepop没引用 -done
% TODO 说明一下我们是从2.3W中sample 8K的,说明一下为什么这样做  -done
\textbf{Training setup.} The full training dataset consists of 23k instances. To balance computational cost and enable efficient ablation, all experiments on Qwen3-4B-Thinking-2507 use a randomly sampled subset of 8k instances; the full dataset is used for Qwen3-30B-A3B-Thinking-2507. All methods on the 8k subset are trained under identical conditions. We use the \texttt{verl} framework in a strictly on-policy setting with a batch size of 128, a group size of 16, and a learning rate of 2e-6 with 5 warmup steps. For rollout generation, temperature and top-p are both set to 1.0 to promote exploration. We additionally adopt IcePop~\citep{team2025every} to mitigate the mismatch between training and inference in long-context settings.

% TODO 这个部分写的稍微有点啰嗦,大概讲一下每个bmk评测的是什么能力,点一下和qwenlong L1.5的评测的bmk是对齐的。-done
\textbf{Evaluation.} We evaluate on a diverse set of long-context benchmarks, following the evaluation protocol of QwenLong-L1.5 for fair comparison (detailed scoring criteria in Appendix~\ref{app:eval_alignment}). The six benchmarks are \textbf{LongBench-V2}~\citep{bai2025longbench} for general long-context comprehension, \textbf{MRCR}~\citep{vodrahalli2024michelangelo} for retrieval under long contexts, \textbf{Frames}~\citep{krishna2025fact} for multi-hop reasoning, \textbf{LongBench}~\citep{bai2024longbench} with five subsets (2WikiMultihopQA~\citep{ho-etal-2020-constructing}, HotpotQA~\citep{yang2018hotpotqa}, MuSiQue~\citep{trivedi2022musique}, NarrativeQA~\citep{kovcisky2018narrativeqa}, and Qasper~\citep{dasigi2021dataset}), \textbf{DocMath}~\citep{zhao2024docmath} for numerical reasoning over documents, and \textbf{CorpusQA}~\citep{lu2026corpusqa} for corpus-level multi-document reasoning.
% TODO CorpusQA这个是不是描述稍微有点不准确

% We further evaluate generalization beyond long-context reasoning following QwenLong-L1.5. Specifically, we assess general capabilities on MMLU-Pro~\citep{wang2024mmlu}, AIME24/25~\citep{aime2025}, and GPQA-Diamond~\citep{rein2023gpqa}, agentic memory using the BFCL-V4 memory subset~\citep{patil2025berkeley}, and dialogue memory using LongMemEval~\citep{wu2024longmemeval}. The results, which demonstrate that our approach maintains or improves performance on these tasks, indicating positive generalization without noticeable forgetting, can be found in Appendix~\ref{app:experiments}.

To verify that long-context RL training does not degrade general capabilities, we further evaluate on standard reasoning and memory benchmarks, including MMLU-Pro~\citep{wang2024mmlu}, AIME24/25~\citep{aime2025}, GPQA-Diamond~\citep{rein2023gpqa}, the BFCL-V4 memory subset~\citep{patil2025berkeley} for agentic memory, and LongMemEval~\citep{wu2024longmemeval} for dialogue memory.

% The results, which demonstrate that our approach maintains or improves performance on these tasks, indicating positive generalization without noticeable forgetting, can be found in Appendix~\ref{app:experiments}.

% TODO 需要加的比较多:
% 1. 说一下我们的评测脚本和qwenlong L1.5是对齐的,包括评测长度,参数等等,评测复现效果见附录
% 2. 简单说明一下我们除了测6个bmk,还参考了qwenlong L1.5的通用评测和超长上下文评测。具体的设置和实验效果看附录

% TODO 我们每个实验都是怎么设置的,这个具体得说明一下
% \textbf{Compared Methods.} We consider several designs of advantage estimation, including GRPO, task-aware advantage estimation (QwenLong-L1.5), and our proposed method. GRPO performs group-level reward normalization for advantage computation. Task-aware advantage estimation incorporates task-level information by normalizing advantages using the reward standard deviation of samples within the same task, mitigating cross-task distributional differences. Our method consists of two components: Task-level Mean Normalization (TMN), which replaces per-prompt normalization with task-level normalization to remove cross-task scale differences, and Difficulty-adaptive Reweighting, which further adjusts advantages according to sample difficulty. We also report results using TMN alone to analyze the contribution of each component. All methods share the same training configuration and differ only in the design of advantage estimation, ensuring fair comparison.

\textbf{Compared Methods.} We compare standard GRPO with our proposed TMN-Reweight. All methods share the same training configuration except for the advantage estimation strategy, ensuring a fair comparison. Detailed training hyperparameters are provided in Appendix~\ref{app:train_config}.

% TODO 这个等待实验结果出来再分析
\subsection{Main Results}

% We evaluate our method on a diverse suite of long-context benchmarks and compare it with Qwen3-4B-Thinking-2507, a vanilla GRPO baseline, and the representative long-context post-training method QwenLong-L1.5. The main results are shown in Table~\ref{tab:main_results}.

The main results are shown in Table~\ref{tab:main_results}.

\begin{table}[t!]
\centering
\begin{threeparttable}
% \caption{Main results on long-context benchmarks.}
\caption{Long-context benchmark results and component ablation. Rows \textit{+GRPO} and \textit{+TMN-Reweight} share the same capability-oriented training data and differ only in the advantage estimation method, isolating the algorithmic contribution.}
\label{tab:main_results}
\footnotesize
\setlength{\tabcolsep}{4pt}
\renewcommand{\arraystretch}{1.1}
\begin{tabular}{lcccccccc}
\toprule
\textbf{Model} & \textbf{Avg.} & \textbf{DocMath} & \textbf{LBV2}
  & \textbf{Frames} & \textbf{MRCR} & \textbf{CorpusQA} & \textbf{LBV1-QA} \\
\midrule
QwenLong-L1.5-4B\tnote{$\dagger$}
  & 59.4 & \textbf{62.5} & \textbf{47.9} & \textbf{67.4} & 47.9 & 64.7 & 65.8 \\
\cmidrule(l){1-8}
Qwen3-4B-Thinking-2507
  & 53.0 & 61.0 & 40.2 & 64.4 & 38.4 & 49.9 & 64.0 \\
\textbf{GoLongRL-4B} \textit{(w. GRPO)}~
  & \underline{62.2} & \textbf{62.5} & 45.5 & 66.6 & \textbf{67.5} & \underline{65.1} & \textbf{65.9} \\
\textbf{GoLongRL-4B} \textit{(w. TMN-Reweight)}
  & \textbf{63.0} & \underline{62.3} & \underline{47.1} & \textbf{67.4} & \underline{65.5} & \textbf{69.6} & \textbf{65.9} \\
\bottomrule
\end{tabular}%
\begin{tablenotes}
  \footnotesize
  \item[$\dagger$] Results reported directly from \citep{qwenlong},
  as the 4B model has not been publicly released.
  All other results are evaluated under our unified evaluation protocol.
\end{tablenotes}
\end{threeparttable}
\end{table}

% Overall, long-context reinforcement learning substantially improves the base model across all evaluated settings. Building upon the GRPO baseline, TMN-Reweight further improves the average score to 63.0, achieving the best overall performance among all compared methods and surpassing QwenLong-L1.5-4B by 3.6 points.

Compared with the vanilla GRPO baseline, TMN-Reweight improves the average score from 62.2 to 63.0, with gains concentrated on aggregation- and reasoning-intensive benchmarks such as CorpusQA (+4.5) and LBV2 (+1.6). On retrieval-oriented tasks such as MRCR, vanilla GRPO achieves the higher score (67.5 vs.\ 65.5), suggesting that the reweighting mechanism is more beneficial when tasks require coordinating multiple reasoning capabilities rather than optimizing a single dominant behavior.

From a per-task perspective, different methods exhibit distinct strength profiles. The vanilla GRPO baseline achieves the highest score on MRCR (67.5), indicating that direct RL optimization is highly effective for retrieval-oriented tasks. QwenLong-L1.5, while competitive on DocMath, LBV2, and Frames, scores only 47.9 on MRCR, revealing a notable weakness in long-context retrieval. In contrast, TMN-Reweight achieves the best or second-best result on five out of six sub-tasks, demonstrating a more balanced capability profile. In particular, on CorpusQA, TMN-Reweight achieves the highest score of 69.6, outperforming the GRPO baseline by 4.5 points and QwenLong-L1.5 by 4.9 points, indicating that the proposed reweighting mechanism is especially effective for tasks requiring corpus-level information aggregation.

These results suggest that the core advantage of TMN-Reweight lies not in maximizing performance on any single task, but in providing more stable and informative training signals that improve robustness and generalization across a diverse set of long-context reasoning tasks ~\citep{DBLP:journals/corr/abs-2512-05591}.
% TODO 这个等待实验结果出来再分析
\subsection{Ablation study}

All TMN-Reweight ablations are conducted at the 4B scale,
where the dense architecture provides a controlled environment
for attributing performance differences to specific algorithmic
modifications.
The 30B MoE model introduces confounding factors such as
train-inference inconsistency in expert routing, which
requires dedicated stabilization mechanisms like R3~\citep{r3}.
We therefore validate TMN-Reweight in the 4B setting and
leave MoE-specific integration to future work.
The data contribution is evaluated separately at both scales
in Section~\ref{sec:data_verify}.

We conduct ablation experiments to analyze the effectiveness of TMN-Reweight and the effect of its key hyperparameter.

% \textbf{Component ablation.} We first evaluate the impact of different components in advantage estimation. Starting from standard GRPO, we progressively introduce the two key modules in our method: Task-level Mean Normalization (TMN) and Difficulty-adaptive Reweighting. Specifically, we consider three settings: (1) GRPO, (2) GRPO + TMN, and (3) GRPO + TMN + Reweighting. This study aims to examine whether TMN effectively mitigates cross-task scale inconsistency and whether the addition of difficulty-adaptive reweighting further improves performance. Results are shown in Table~\ref{tab:ablation_components}. % table to be added

% \textbf{Component ablation.} Table~\ref{tab:main_results} also serves as a component ablation, since the \textit{+GRPO} and \textit{+TMN-Reweight} rows share identical training data and differ only in advantage estimation. The gap between the two rows is 0.8 points on average (62.2 vs.\ 63.0), suggesting that TMN-Reweight provides a modest but consistent improvement over vanilla GRPO on this dataset. The larger gain comes from the data itself: as shown in Section~\ref{sec:data_verify}, vanilla GRPO trained on our capability-oriented data already outperforms QwenLong-L1.5 with GRPO by 6.1 points, indicating that data coverage is the primary driver of improvement. The algorithmic gain from 
% TMN-Reweight is concentrated on aggregation-intensive benchmarks such as CorpusQA (+4.5) and LBV2 (+1.6), while retrieval-oriented tasks such as MRCR slightly favor vanilla GRPO.

\textbf{Component ablation.} Table~\ref{tab:main_results} also serves as a component ablation, since the \textit{+GRPO} and \textit{+TMN-Reweight} rows share identical training data and differ only in advantage estimation. The 0.8-point average gap (62.2 vs.\ 63.0) indicates that TMN-Reweight provides a modest but consistent gain over vanilla GRPO. The larger driver is the data itself, as vanilla GRPO on our data already outperforms QwenLong-L1.5 with GRPO by 6.1 points (Section~\ref{sec:data_verify}). The algorithmic contribution of TMN-Reweight is most pronounced on aggregation-intensive benchmarks, with CorpusQA improving by +4.5 points and LBV2 by +1.6 points, while retrieval-oriented tasks such as MRCR slightly favor vanilla GRPO. The pronounced CorpusQA improvement is consistent with the design intent of TMN-Reweight, which targets settings where heterogeneous reward signals must be jointly balanced.
 
\begin{table}[t!]
\centering
\caption{Effect of $\alpha$ in difficulty-adaptive reweighting. Best results per column are \textbf{bolded}, second best are \underline{underlined}.}
\label{tab:alpha_ablation}
\footnotesize
\setlength{\tabcolsep}{5pt}
\begin{tabular}{c c cccccc}
\toprule
$\alpha$ & \textbf{Avg.} & \textbf{DocMath} & \textbf{LBV2} & \textbf{Frames} & \textbf{MRCR} & \textbf{CorpusQA} & \textbf{LBV1-QA} \\
\midrule
0.0 & 61.3 & 61.9 & 45.8 & \textbf{68.7} & 64.5 & 60.2 & \textbf{66.5} \\
0.5 & \underline{62.9} & \textbf{63.5} & \textbf{47.5} & \underline{67.4} & \underline{64.9} & \underline{67.8} & \underline{66.2} \\
0.8 & \textbf{63.0} & \underline{62.4} & \underline{47.1} & 67.4 & \textbf{65.5} & \textbf{69.6} & 65.9 \\
1.0 & 61.5 & 62.4 & 44.9 & 66.8 & 65.4 & 65.7 & 63.8 \\
\bottomrule
\end{tabular} 
\end{table}

% TODO alpha参数消融实验
\textbf{Effect of $\alpha$.} We further investigate the effect of the hyperparameter $\alpha$ in Difficulty-adaptive Reweighting. This parameter controls the trade-off between prompt-level and task-level difficulty estimation. Larger $\alpha$ relies more on prompt-level statistics but may introduce higher variance, while smaller $\alpha$ leads to more stable estimates at the cost of reduced granularity. To study this trade-off, we vary $\alpha$ while keeping other settings fixed.

Results are shown in Table~\ref{tab:alpha_ablation}. The best average performance is achieved at $\alpha=0.8$, outperforming both extremes of $\alpha=1.0$ (pure prompt-level, 61.5) and $\alpha=0.0$ (pure task-level, 61.3). Prompt-level estimation alone introduces high variance with small group sizes, while task-level estimation alone loses prompt-specific difficulty information. The intermediate value $\alpha=0.8$ retains most of the prompt-level granularity while borrowing sufficient stability from the task-level baseline, and is used as the default in all other experiments.

\subsection{Analysis}

% TODO 放一下我们模型在通用榜单上的效果
\textbf{General Capability Retention}
To evaluate the generalization of skills acquired through long-context post-training, we conduct evaluations on both \textbf{Qwen3-4B-Thinking-2507} and \textbf{Qwen3-30B-A3B-Thinking-2507} across three dimensions: general reasoning, agentic memory, and dialogue memory (Table~\ref{tab:general_results}).

% \begin{table}[t!]
% \centering
% \caption{Results on general reasoning, agentic memory, and dialogue memory benchmarks.}
% \label{tab:general_results}
% \footnotesize
% \setlength{\tabcolsep}{5pt}
% \begin{tabular}{l *{4}{c}}
% \toprule
% \multirow{2}{*}{\textbf{Benchmark}}
%   & \multicolumn{2}{c}{\textbf{Qwen3-4B-Thinking-2507}}
%   & \multicolumn{2}{c}{\textbf{Qwen3-30B-A3B-Thinking-2507}} \\
% \cmidrule(lr){2-3} \cmidrule(lr){4-5}
%   & Base & +\,TMN-Reweight
%   & Base & +\,TMN-Reweight \\
% \midrule
% \multicolumn{5}{c}{\cellcolor{gray!8}\textit{General Reasoning}} \\[2pt]
% MMLU-Pro        & 73.6 & 74.3\up{0.7}   & 80.2  & 81.2\up{1.0} \\
% AIME24          & 82.6 & 84.2\up{1.6}   & 90.5  & 89.8\down{0.7} \\
% AIME25          & 80.2 & 80.8\up{0.6}   & 85.1  & 86.6\up{1.5} \\
% GPQA            & 65.4 & 67.7\up{2.3}   & 70.1  & 71.0\up{0.9} \\[3pt]
% \multicolumn{5}{c}{\cellcolor{gray!8}\textit{Agentic Memory}} \\[2pt]
% Memory-KV       & 12.9 & 11.6\down{1.3} & 15.5  & 12.9\down{2.6} \\
% Memory-Vec      & 15.5 & 20.0\up{4.5}   & 18.7  & 16.8\down{1.9} \\
% Memory-Rec\_Sum  & 36.8 & 46.5\up{9.7}  & 31.0  & 40.0\up{9.0} \\[3pt]
% \multicolumn{5}{c}{\cellcolor{gray!8}\textit{Dialogue Memory}} \\[2pt]
% LongMemEval     & 47.6 & 61.2\up{13.6}  & 61.6  & 71.2\up{9.6} \\
% \bottomrule
% \end{tabular}
% \end{table}

\begin{table}[t!]
\centering
\caption{Results on general reasoning, agentic memory, and dialogue memory benchmarks.}
\label{tab:general_results}
\footnotesize
\setlength{\tabcolsep}{5pt}
\begin{tabular}{l cc | c cc}
\toprule
\multirow{2}{*}{\textbf{Benchmark}}
  & \multicolumn{2}{c|}{\textbf{Qwen3-4B-Thinking-2507}}
  & \textbf{QwenLong}
  & \multicolumn{2}{c}{\textbf{Qwen3-30B-A3B-Thinking-2507}} \\
\cmidrule(lr){2-3} \cmidrule(lr){5-6}
  & Base & GoLongRL (w. TMN-Reweight)
  & \textbf{L1.5-30B}
  & Base & GoLongRL (w. GRPO) \\
\midrule
\multicolumn{6}{c}{\cellcolor{gray!8}\textit{General Reasoning}} \\
\midrule
MMLU-Pro        & 73.6 & 74.3\up{0.7}   & 81.1  & 80.2 & 81.0\up{0.8}   \\
AIME24          & 82.6 & 84.2\up{1.6}   & 89.8  & 90.5 & 91.3\up{0.8}   \\
AIME25          & 80.2 & 80.8\up{0.6}   & 87.9  & 85.1 & 86.9\up{1.8}   \\
GPQA-Diamond    & 65.4 & 67.7\up{2.3}   & 72.6  & 70.1 & 72.3\up{2.2}   \\
\midrule
\multicolumn{6}{c}{\cellcolor{gray!8}\textit{Agentic Memory}} \\
\midrule
Memory-KV       & 12.9 & 11.6\down{1.3} & \textbf{16.1} & 15.5 & \textbf{16.1}\up{0.6}   \\
Memory-Vec      & 15.5 & 20.0\up{4.5}   & 21.3  & 18.7 & \textbf{21.9}\up{3.2}   \\
Memory-Rec\_Sum & 36.8 & 46.5\up{9.7}   & \textbf{36.1} & 31.0 & 35.5\up{4.5}   \\
\midrule
\multicolumn{6}{c}{\cellcolor{gray!8}\textit{Dialogue Memory}} \\
\midrule
LongMemEval     & 47.6 & 61.2\up{13.6}  & 72.2  & 61.6 & \textbf{75.2}\up{13.6}  \\
\bottomrule
\end{tabular}
\end{table}

In terms of general reasoning, TMN-Reweight yields improvements on all four benchmarks at both scales. The 4B model gains on MMLU-Pro (+0.7), AIME24 (+1.6), AIME25 (+0.6), and GPQA-Diamond (+2.3), with a similar trend at 30B. These results suggest that information integration skills learned during long-context training transfer effectively to standard reasoning tasks.

% In the agentic memory tasks, the 4B model shows a slight decline on Memory-KV ($-$1.3), potentially due to the sensitivity of precise key-value retrieval, while Memory-Vec and Memory-Rec\_Sum achieve substantial improvements of +4.5 and +9.7, respectively. For the 30B model, all three agentic memory benchmarks improve: Memory-KV (+0.6), Memory-Vec (+3.2), and Memory-Rec\_Sum (+4.5). This indicates that long-context training provides stable gains in semantic-level information integration across both model scales.

% In the agentic memory tasks, the 4B model shows a slight decline on Memory-KV ($-$1.3), potentially due to the sensitivity of precise key-value retrieval, while Memory-Vec and Memory-Rec\_Sum improve substantially (+4.5 and +9.7, respectively). The 30B model shows consistent gains across all three agentic benchmarks: Memory-KV (+0.6), Memory-Vec (+3.2), and Memory-Rec\_Sum (+4.5). Together, these results suggest that long-context RL training transfers effectively to semantic retrieval and state-tracking tasks, with more reliable gains at the 30B scale.

For agentic memory, both models show overall improvements, with notable gains on Memory-Vec and Memory-Rec\_Sum, indicating that long-context RL training transfers to semantic retrieval and state-tracking tasks. The 4B model shows a marginal decline on Memory-KV ($-$1.3), while the 30B model improves consistently across all three subtasks.

For dialogue memory, LongMemEval evaluates a model's ability to maintain state and recall information across extended conversations. Both the 4B and 30B models achieve substantial improvements of +13.6, demonstrating significant gains in long-range dependency handling.

In summary, long-context training with TMN-Reweight preserves general reasoning capabilities while achieving substantial improvements on information integration and long-range context tasks at both scales.

% TODO 放一下我们超长bmk的表现
% 长度外推时候的配置
\textbf{Length Extrapolation Performance.}
We evaluate the extrapolation capability of GoLongRL on long-sequence MRCR and CorpusQA tasks (Table~\ref{tab:ablation}). Although the model was trained with a context length of 160K, its long-context capability generalizes effectively to longer evaluation sequences.

On the 4B model, GoLongRL (GRPO) achieves gains of +12.27 on MRCR 128K--512K and +3.50 on 512K--1M, with TMN-Reweight showing a similar trend. The 30B model shows even larger gains, improving by +12.61 on MRCR 128K--512K, +5.45 on MRCR 512K--1M, and +2.74 on CorpusQA 1M. These results suggest that information integration skills learned during 160K training extrapolate to much longer sequences, consistently across model scales.
\begin{table}[t!]
\centering
\caption{Ablation results on long-context benchmarks (128K+). Best results per section are \textbf{bolded}.}
\label{tab:ablation}
\footnotesize
\setlength{\tabcolsep}{5pt}
\begin{tabular}{l *{3}{c}}
\toprule
\multirow{2}{*}{\textbf{Models}}
  & \multicolumn{2}{c}{\textbf{MRCR}}
  & \textbf{CorpusQA} \\
\cmidrule(lr){2-3} \cmidrule(lr){4-4}
  & 128K--512K & 512K--1M & 1M \\
\midrule
\multicolumn{4}{c}{\cellcolor{gray!8}\textit{Qwen3-4B}} \\
\midrule
Qwen3-4B-Thinking-2507                    & 11.16 & 1.37 & \textbf{5.78} \\
GoLongRL (w. GRPO)                   & \textbf{23.43} & \textbf{4.87} & 5.47 \\
GoLongRL (w. TMN-Reweight)          & 21.85 & 3.87 & 4.56 \\[3pt]
\midrule
\multicolumn{4}{c}{\cellcolor{gray!8}\textit{Qwen3-30B-A3B}} \\
\midrule
Qwen3-30B-A3B-Thinking-2507                   & 24.91   & 3.69  & 6.38  \\
GoLongRL (w. GRPO)              & \textbf{37.52}   & \textbf{9.14}  & \textbf{9.12}  \\
\bottomrule
\end{tabular}
\end{table}

% TODO 可选：可以做一下delta ppl的实验分析：
% 计算四组PPL
% PPL_gold_with_ctx    = PPL(gold | c, q)
% PPL_gold_no_ctx      = PPL(gold | q)
% PPL_resp_with_ctx    = PPL(resp | c, q)   # 按correct/incorrect分层
% PPL_resp_no_ctx      = PPL(resp | q)

% 期望结论
% |  | **Δ PPL 大** | **Δ PPL 小** |
% |---|---|---|
% | **Correct** | 理想情况：真正学会了用上下文 | 模型靠 prior 猜对了 |
% | **Incorrect** | 模型在幻觉但仍依赖上下文 | 完全没利用上下文 |

% RL训练过程中，Δ PPL单调上升（上下文依赖增强）
% Correct responses的Δ PPL >> incorrect responses的Δ PPL（依赖上下文 → 答对）
% Gold answer的Δ PPL变化可以作为independent verification（排除response distribution shift的影响）

% % TODO DW-HA这个文章没有通过实验评估系统性高估、低估了多少，我们可以用rollout 48次，然后从中随机选取16个（模拟RL过程，全零全1直接排除），看看系统性高估或低估了多少
% \textbf{Systemic advantage prediction bias analysis}

% % TODO 不同实验训练时的曲线展示和分析

%% file: sections/conclusion.tex
\section{Conclusion and Future Work}
\label{sec:conclusion}

% We present a capability-oriented long-context RL framework that mitigates two limitations of existing approaches, namely narrow data coverage and unstable multitask optimization under heterogeneous rewards. The dataset contains 23K samples across 9 task types, combining curated open-source data with synthetic samples grounded in real documents and refined through quality control. TMN-Reweight combines task-level reward scale normalization with difficulty-adaptive reweighting, yielding improved average performance at the 4B scale. On Qwen3-4B-Thinking, the capability-oriented dataset with vanilla GRPO raises the long-context benchmark average from 53.0 to 62.2, and TMN-Reweight further improves it to 63.0, surpassing QwenLong-L1.5 in the same setting. The reported general capability evaluations suggest that these gains do not come at the cost of broad reasoning or memory degradation. The dataset, data pipeline, and training code are open-sourced.

We present a capability-oriented long-context RL framework that mitigates two limitations of existing approaches, namely narrow data coverage and unstable multitask optimization under heterogeneous rewards. The dataset contains 23K samples across 9 task types, combining curated open-source samples with synthetic samples whose QA pairs are generated from real documents. Both tracks are produced through a four-phase construction pipeline with multi-stage quality control and iterative refinement. TMN-Reweight combines task-level reward scale normalization with difficulty-adaptive reweighting, yielding improved average performance at the 4B scale. On Qwen3-4B-Thinking, the capability-oriented dataset with vanilla GRPO raises the long-context benchmark average from 53.0 to 62.2, and TMN-Reweight further improves it to 63.0, surpassing QwenLong-L1.5. General capability evaluations suggest that these gains do not come at the cost of reasoning or memory degradation. The dataset, data pipeline, and training code are open-sourced.

\label{sec:future}
\textbf{Future work.} The effectiveness of difficulty reweighting may vary across model scales, with clearer gains observed at 4B and less settled evidence at larger scales. This motivates further study of scale-dependent optimization dynamics. The remaining CorpusQA gap suggests that targeted data supplementation for multi-document reasoning may be beneficial. Additionally, combining the context-aware token weighting method (Section~\ref{sec:token_reweighting}) with RLVR for more fine-grained training would be a promising research direction.

%% file: sections/appendix.tex
% =============================================================
%  Appendix
% =============================================================

% Prompt box style is defined in main.tex
% 
% The original tcbset configuration from appendix__2_.tex has been commented out:
% \tcbset{
%   promptbox/.style={
%     colback=codebg,
%     colframe=codeframe,
%     fonttitle=\bfseries\small,
%     title={#1},
%     boxrule=0.5pt,
%     arc=2pt,
%     left=4pt, right=4pt, top=4pt, bottom=4pt,
%     breakable,
%     enhanced,
%   }
% }

\section{Gradient Analysis and Theoretical Motivation for TMN}
\label{app:tmn_theory}

This appendix provides a gradient-based derivation that motivates the task-level normalization in TMN (Section~\ref{sec:tmn_reweight}). The analysis starts from the REINFORCE-style policy gradient used by Dr.\ GRPO and shows that the per-task gradient magnitude is governed by the expected within-prompt reward variance. Equalizing these magnitudes across tasks leads naturally to the $\sigma_{\text{task}}$ denominator introduced in the main text.

\subsection{Per-Prompt Gradient Norm Bound}

Consider a GRPO variant that removes the per-prompt standard deviation, as in Dr.\ GRPO. The policy gradient takes the REINFORCE form
\begin{equation}
\label{eq:app_pg}
    \nabla_\theta \mathcal{J} = \mathbb{E}_{u,\,o}\!\Big[\big(r(u,o) - \mu_u\big)\,\nabla_\theta \log \pi_\theta(o \mid u)\Big]
\end{equation}
where $u$ denotes the prompt, $o$ the sampled response, $r(u,o)$ the reward, and $\mu_u = \mathbb{E}_{o}[r(u,o)]$ the expected reward under prompt $u$. For a single prompt $u$, the per-prompt gradient is
\begin{equation}
    \nabla_\theta \mathcal{J}_u = \mathbb{E}_{o}\!\Big[\big(r(u,o) - \mu_u\big)\,\nabla_\theta \log \pi_\theta(o \mid u)\Big]
\end{equation}
By the Cauchy--Schwarz inequality applied inside the expectation,
\begin{equation}
\label{eq:app_bound}
    \big\|\nabla_\theta \mathcal{J}_u\big\|^2 \;\leq\; \mathbb{E}_{o}\!\Big[\big(r(u,o) - \mu_u\big)^2 \;\big\|\nabla_\theta \log \pi_\theta(o \mid u)\big\|^2\Big]
\end{equation}
Assuming that the log-probability gradient norm $\|\nabla_\theta \log \pi_\theta(o \mid u)\|^2$ varies modestly compared with the reward deviation $(r - \mu_u)^2$, the dominant factor controlling the per-prompt gradient magnitude is the within-prompt reward variance
\begin{equation}
    \text{Var}(r \mid u) = \mathbb{E}_{o}\!\Big[\big(r(u,o) - \mu_u\big)^2\Big] = \sigma_u^2
\end{equation}
This quantity is exactly the square of the per-prompt standard deviation $\sigma_u$ defined in Eq.~\ref{eq:grpo_adv} of the main text.

\subsection{Per-Task Gradient Magnitude}

Now consider $K$ task types $\{\mathcal{T}_k\}_{k=1}^{K}$, each with its own reward function $r^{(k)}$. The overall gradient decomposes as
\begin{equation}
    \nabla_\theta \mathcal{J} = \mathbb{E}_{k \sim \mathcal{P}(\mathcal{T})}\;\mathbb{E}_{u \sim \mathcal{T}_k,\, o}\!\Big[\big(r^{(k)}(u,o) - \mu_u\big)\,\nabla_\theta \log \pi_\theta(o \mid u)\Big]
\end{equation}
The gradient contribution from task $k$ is
\begin{equation}
    \nabla_\theta \mathcal{J}_k = \mathbb{E}_{u \sim \mathcal{T}_k,\, o}\!\Big[\big(r^{(k)}(u,o) - \mu_u\big)\,\nabla_\theta \log \pi_\theta(o \mid u)\Big]
\end{equation}
Extending the per-prompt analysis, the magnitude of $\nabla_\theta \mathcal{J}_k$ is governed by the expected within-prompt reward variance across prompts in task $k$,
\begin{equation}
\label{eq:app_task_var}
    \mathbb{E}_{u \sim \mathcal{T}_k}\!\big[\sigma_u^2\big] = \mathbb{E}_{u \sim \mathcal{T}_k}\!\big[\text{Var}(r^{(k)} \mid u)\big]
\end{equation}
Tasks whose reward functions produce higher variance, such as continuous-valued F1 or NDCG metrics, will contribute larger gradient norms than tasks with lower-variance rewards, such as EM or multiple-choice Accuracy (both binary-valued). Without explicit normalization, the former can dominate the joint optimization.

\subsection{Deriving the Normalization Coefficient}

To equalize gradient magnitudes across tasks, we introduce a per-task scaling coefficient $\alpha_k$ that normalizes the advantage by the expected gradient scale. Setting the target to unit gradient magnitude yields
\begin{equation}
\label{eq:app_alpha}
    \alpha_k = \frac{1}{\sqrt{\mathbb{E}_{u \sim \mathcal{T}_k}[\sigma_u^2]} + \delta}
\end{equation}
The empirical estimate of $\mathbb{E}_{u \sim \mathcal{T}_k}[\sigma_u^2]$ over a batch containing $|U_{\text{task}}|$ prompts from task $k$ is
\begin{equation}
    \hat{\mathbb{E}}[\sigma_u^2] = \frac{1}{|U_{\text{task}}|}\sum_{u \in U_{\text{task}}} \sigma_u^2
\end{equation}
Taking the square root recovers $\sigma_{\text{task}(i)}$ as defined in Eq.~5 of the main text,
\begin{equation}
    \sigma_{\text{task}(i)} = \sqrt{\frac{1}{|U_{\text{task}}|}\sum_{u \in U_{\text{task}}} \sigma_u^2}
\end{equation}
This establishes that the TMN denominator is the natural normalization constant for equalizing per-task gradient contributions under the variance-based gradient bound in Eq.~\ref{eq:app_bound}.

The derivation above provides the theoretical basis for the $\sigma_{\text{task}(i)}$ formula in Eq.~5 of Section~\ref{sec:tmn_reweight}. \textbf{The per-task gradient magnitude is governed by $\mathbb{E}_{u \sim \mathcal{T}_k}[\sigma_u^2]$, which is the expected within-prompt reward variance, not by $\text{Var}_{u,o}(r^{(k)})$, which is the total variance obtained by pooling all reward samples across prompts.} These two quantities differ because the pooled variance conflates within-prompt variation (which drives gradient magnitude) with between-prompt variation in mean reward (which reflects difficulty differences rather than scale differences). Computing $\sigma_{\text{task}(i)}$ as the root mean square of per-prompt $\sigma_u$ values estimates precisely the former, yielding a normalization constant that equalizes gradient scales without distorting the difficulty structure that TMN-Reweight subsequently exploits in its second step.

\subsection{Structural Relationship to GRPO}

The task-level coefficient $\alpha_k$ in Eq.~\ref{eq:app_alpha} bears a notable structural resemblance to the per-prompt normalization used by standard GRPO. In GRPO, the advantage for each response is divided by $\sigma_u$, which is equivalent to applying a prompt-level coefficient
\begin{equation}
    \eta_u = \frac{1}{\sigma_u + \delta}
\end{equation}
Both $\alpha_k$ and $\eta_u$ normalize by a standard deviation to balance reward scales, and both can be viewed as instances of the same variance-based principle. The critical difference lies in the granularity at which normalization is applied. GRPO applies $\eta_u$ at the prompt level, which addresses cross-task scale differences but simultaneously distorts within-task difficulty structure, because prompts with small $\sigma_u$ (easy or hard) receive inflated advantages. TMN applies $\alpha_k$ at the task level, which addresses cross-task scale differences through a shared denominator while preserving relative advantage magnitudes within each task, leaving the difficulty structure intact for subsequent reweighting.

This analysis clarifies the empirical observation in Figure~\ref{fig:advantage_norms}. In single-task settings with a uniform reward function, removing $\sigma_u$ entirely (as in Dr.\ GRPO) often improves performance because the per-prompt normalization introduces difficulty bias without providing useful scale correction. In our multitask setting, however, removing $\sigma_u$ exposes the gradient to uncorrected cross-task scale differences, which can degrade performance. TMN resolves this tension by retaining normalization at the task level, where it serves the useful function of scale alignment, while removing it at the prompt level (within each task), where it would distort difficulty signals. The result is a method that inherits the scale-balancing property of GRPO across tasks and the unbiased advantage estimation of REINFORCE within each task.

\section{Data Generation Prompts}
\label{app:prompts}

This appendix documents the prompts used to generate training samples in the capability-oriented dataset described in Section~\ref{sec:data}. Each prompt is organized under the corresponding task in our 9-task taxonomy. Within each task, subtask prompts specify the context requirement (Full or Partial), the evaluation metric, and the expected output format. All prompts follow a unified structure: a primary task definition establishes the capability being assessed, while secondary task definitions specify the concrete subtask, its scope, and the answer format.

The sample generation pipeline feeds each prompt together with a long source document into the synthesis model. The model then produces candidate questions and reference answers, which are subsequently parsed, validated through QA-pair verification by Gemini-2.5-Pro, and quality-calibrated through multi-stage verification as described in Section~\ref{sec:data_pipeline}.

% ----- Shared Generation Instruction -----
\subsection{Shared Generation Instruction}
\label{app:shared_prompt}

All task-specific prompts are wrapped inside the following shared instruction template, which guides the synthesis model through language identification, question construction, identifier design, and self-checking.

\begin{promptbox}[Shared Sample Generation Instruction]
\begin{lstlisting}[style=promptstyle]
You are a professional question-design specialist with capabilities in textual deconstruction, deep reading, and cross-domain item construction.

Task: Given an input long context and its task category (primary task + secondary task), create three questions that fulfill the task requirements, including answers, design rationale, and detailed solution process.

Procedure:
1. Language Identification
   Determine the language of the input long text. All subsequent content must fully use the same language.

2. Question Construction
   - Generate three questions satisfying the secondary task specifications.
   - Questions must meet the evaluation goals of the primary task, the required I/O format, and the secondary task definition.
   - The three questions must exhibit clear differentiation, avoiding repeated formats or duplicate assessment points.
   - Identifier Rules: Use naturally occurring unique elements (titles, section names, entities) as identifiers. If insufficient, construct identifiers using only letters and numbers.
   - For multiple-choice questions: include plausible distractors, at least four options, and a text-verifiable correct option.
   - Provide design rationale and solution with evidence for every question.

3. Self-Check
   Verify task compliance, answer accuracy, absence of ambiguity, and language consistency. Regenerate if any criterion fails.

Input:
[Long Context]: {Long_Context}
[Primary Task]: {Primary_Task}
[Secondary Task]: {Secondary_Task}
\end{lstlisting}
\end{promptbox}

\subsection{QA-Pair Verification Prompt}
\label{app:verification_prompt}

After QA generation, each sample undergoes QA-pair verification. The following prompt is sent to Gemini-2.5-Pro together with the full source document and the generated QA pair. Samples that receive a REJECT verdict are discarded before multi-stage verification.

\begin{promptbox}[QA-Pair Verification Prompt]
\begin{lstlisting}[style=promptstyle]
You are a rigorous quality assessor for long-context question-answer pairs. Given a source document and a generated QA sample (question, answer options, the designated correct answer, and the solution rationale), perform all of the following checks.

1. Answer Uniqueness
   Verify that the correct answer is uniquely and unambiguously supported by the source document. The answer must be derivable from the text without requiring external knowledge or subjective interpretation. If the document could support more than one option, flag the sample.

2. Distractor Quality
   Confirm that each distractor (incorrect option) is plausible given the topic but clearly wrong according to the source document. Flag any distractor that is trivially distinguishable, or that could also be considered correct under a reasonable reading.

3. Hallucination Detection
   Check whether the question, the answer, or the rationale contains any claim not grounded in the source document. Flag fabricated entities, events, numbers, or causal relationships.

4. Task Compliance
   Verify that the question targets the intended reasoning capability (e.g., cross-passage evidence integration, rule induction, or speaker tracking in dialogue) and that the answer format matches the task specification.

5. Language and Format Consistency
   Confirm that the question and all options are in the same language as the source document and that the output follows the required structured format.

Output format:
Verdict -- PASS or REJECT
If REJECT, specify which check(s) failed and provide a brief explanation for each failure.

Input:
[Source Document]: {Source_Document}
[Question]: {Question}
[Options]: {Options}
[Correct Answer]: {Correct_Answer}
[Solution Rationale]: {Solution_Rationale}
\end{lstlisting}
\end{promptbox}

% =========================================================
%  T1 -- Precise Long-Range Information Retrieval (EM)
% =========================================================
\subsection{T1: Precise Long-Range Information Retrieval}
\label{app:prompt_t1}

T1 evaluates a model's ability to locate and extract specific information across long documents. The reward function is Exact Match (EM). Training data for this task primarily consists of needle-in-a-haystack samples and curated open-source retrieval instances, which do not require task-specific generation prompts beyond the shared instruction above.

% =========================================================
%  T2 -- Evidence-Grounded Comprehension and Reasoning (Accuracy)
% =========================================================
\subsection{T2: Evidence-Grounded Comprehension and Reasoning}
\label{app:prompt_t2}

T2 assesses the model's ability to answer fact-based and reasoning questions grounded in textual evidence. The reward function is multiple-choice Accuracy. This task encompasses multi-hop integration QA, single-hop fact QA, dialogue memory tracking, and related comprehension patterns. Below we present the subtask prompts.

\subsubsection{T2-a: Multi-Document Integration QA}

Answer questions that require synthesizing information from multiple scattered passages across the full document.

\begin{promptbox}[T2-a: Multi-Document Integration QA]
\begin{lstlisting}[style=promptstyle]
Primary Task: Evidence-Grounded QA
Answer fact/reasoning questions based on evidence.

Secondary Task: Multi-Doc Integration QA
Use multi-hop information across documents to answer questions.
Context Requirement: Full
Metric: Accuracy

Output format:
Output the "[Answer]" identifier first, then output the answer option letter (A/B/C/D), without any additional content.

[Answer]
C
\end{lstlisting}
\end{promptbox}

\subsubsection{T2-b: Single-Hop Fact QA}

Answer questions that can be resolved from a localized paragraph without cross-document reasoning.

\begin{promptbox}[T2-b: Single-Hop Fact QA]
\begin{lstlisting}[style=promptstyle]
Primary Task: Evidence-Grounded QA
Answer fact/reasoning questions based on evidence.

Secondary Task: Single-Hop Fact QA
Answer questions based on local paragraphs.
Context Requirement: Partial
Metric: Accuracy

Output format:
Output the "[Answer]" identifier first, then output the answer option letter (A/B/C/D), without any additional content.

[Answer]
A
\end{lstlisting}
\end{promptbox}

\subsubsection{T2-c: Long-Range Entity and Commitment Tracking}

Track entity states and commitments across extended dialogue or narrative context.

\begin{promptbox}[T2-c: Long-Range Entity and Commitment Tracking]
\begin{lstlisting}[style=promptstyle]
Primary Task: Dialogue Memory and Long-Horizon Tracking
Track and respond to dialogue history.

Secondary Task: Long-Range Entity and Commitment Tracking
Track entity states across the global context.
Context Requirement: Full
Metric: Accuracy

Output format:
Output the "[Answer]" identifier first, then output A/B/C/D/E, without any additional content.

[Answer]
B
\end{lstlisting}
\end{promptbox}

\subsubsection{T2-d: Short-Range Reference Resolution}

Resolve references and query local states within a bounded dialogue or narrative segment.

\begin{promptbox}[T2-d: Short-Range Reference Resolution]
\begin{lstlisting}[style=promptstyle]
Primary Task: Dialogue Memory and Long-Horizon Tracking
Track and respond to dialogue history.

Secondary Task: Short-Range Reference Resolution
Resolve references and states in local context.
Context Requirement: Partial
Metric: Accuracy

Output format:
Output the "[Answer]" identifier first, then output the answer option letter (A/B/C/D), without any additional content.

[Answer]
B
\end{lstlisting}
\end{promptbox}

% =========================================================
%  T3 -- High-Recall Exhaustive Retrieval and Verification (F1)
% =========================================================
\subsection{T3: High-Recall Exhaustive Retrieval and Verification}
\label{app:prompt_t3}

T3 targets the model's ability to exhaustively retrieve, verify, and align information with high recall. The reward function is token-level F1. This task aggregates subtasks involving citation alignment, subset identification, compliance checking, and version diff analysis.

\subsubsection{T3-a: Full-Sentence Citation Alignment}

Identify all source locations that support a given summary sentence.

\begin{promptbox}[T3-a: Full-Sentence Citation Alignment]
\begin{lstlisting}[style=promptstyle]
Primary Task: Attribution and Citation Alignment
Bind correct sources to generated text by identifying citation locations.

Secondary Task: Full-Sentence Citation Alignment
Citation alignment for all sentences.
Context Requirement: Full
Metric: F1

Instruction:
You will be provided with a summary sentence. Identify the original Part number(s) from the provided text that fully support this sentence.

Output format:
Output the "[Answer]" identifier first, then output the cited content (format as "Part xx") line by line, without any additional content.

[Answer]
Part 1
Part 12
Part 15
\end{lstlisting}
\end{promptbox}

\subsubsection{T3-b: Key-Statement Citation Alignment}

Identify source locations for a specified sentence within a multi-sentence summary.

\begin{promptbox}[T3-b: Key-Statement Citation Alignment]
\begin{lstlisting}[style=promptstyle]
Primary Task: Attribution and Citation Alignment
Bind correct sources to generated text by identifying citation locations.

Secondary Task: Key-Statement Citation Alignment
Citation alignment for specified sentences.
Context Requirement: Partial
Metric: F1

Instruction:
You will see a generated multi-sentence summary. Identify and cite the original source location for the specified sentence only, using paragraph identifiers from the source text.

Output format:
Output the "[Answer]" identifier first, then output the paragraph identifier(s) line by line, without any additional content.

[Answer]
S1
S2
\end{lstlisting}
\end{promptbox}

\subsubsection{T3-c: Targeted Subset Identification}

Identify instances belonging to a specified category within a collection.

\begin{promptbox}[T3-c: Targeted Subset Identification]
\begin{lstlisting}[style=promptstyle]
Primary Task: Aggregation and Clustering
Cluster and output statistics, examples, or sorted results.

Secondary Task: Targeted Subset Cluster Identification
Return instances matching a query category.
Context Requirement: Partial
Metric: F1

Output format:
Output the "[Answer]" identifier first, then output the IDs that meet the conditions line by line, without any additional content.

[Answer]
A22V1MD93T2FW9
ACJT8MUCOLRFO
A38NELQT98S4H8
\end{lstlisting}
\end{promptbox}

\subsubsection{T3-d: Global Conflict and Inconsistency Localization}

Detect and locate contradictory or inconsistent content across the full text.

\begin{promptbox}[T3-d: Global Conflict and Inconsistency Localization]
\begin{lstlisting}[style=promptstyle]
Primary Task: Consistency and Compliance Checking
Detect and locate contradictions, violations, or conflicts in documents.

Secondary Task: Global Conflict and Inconsistency Localization
Locate contradictory segments in the full text.
Context Requirement: Full
Metric: F1

Output format:
Output the "[Answer]" identifier first, then output all inconsistent article pairs line by line in the format "A[id] B[id]", without any additional content.

[Answer]
A1 B2
A2 B3
\end{lstlisting}
\end{promptbox}

\subsubsection{T3-e: Targeted Rule Violation Detection}

Identify content that violates specific rules or conditions within a designated scope.

\begin{promptbox}[T3-e: Targeted Rule Violation Detection]
\begin{lstlisting}[style=promptstyle]
Primary Task: Consistency and Compliance Checking
Detect and locate contradictions, violations, or conflicts in documents.

Secondary Task: Targeted Rule or Condition Violation Detection
Locate content that violates specific rules.
Context Requirement: Partial
Metric: F1

Output format:
Output the "[Answer]" identifier first, then output all violated term IDs line by line, without any additional content.

[Answer]
5
6
\end{lstlisting}
\end{promptbox}

\subsubsection{T3-f: Comprehensive Error and Anomaly Sweep}

Locate all errors or anomalies (e.g., misspellings) across the full document.

\begin{promptbox}[T3-f: Comprehensive Error and Anomaly Sweep]
\begin{lstlisting}[style=promptstyle]
Primary Task: Consistency and Compliance Checking
Detect and locate contradictions, violations, or conflicts in documents.

Secondary Task: Comprehensive Error and Anomaly Sweep
Locate errors (e.g., spelling) in the full text.
Context Requirement: Full
Metric: F1

Output format:
Output the "[Answer]" identifier first, then output all identified errors line by line, without any additional content.

[Answer]
beleive
freind
tommorow
\end{lstlisting}
\end{promptbox}

\subsubsection{T3-g: Dependency-Aware Multi-Version Impact Analysis}

Track dependency changes across multiple versions of a document or codebase.

\begin{promptbox}[T3-g: Multi-Version Impact Analysis]
\begin{lstlisting}[style=promptstyle]
Primary Task: Version and Code Diff Analysis
Compare changes in different text/code versions.

Secondary Task: Dependency-Aware Multi-Version Impact Analysis
Track dependency changes across versions.
Context Requirement: Full
Metric: F1

Instruction:
Identify all methods/elements that changed status between versions. Use the signature as the identifier in the format "MethodName(ParameterTypes)".

Output format:
Output the "[Answer]" identifier first, then output the identifiers line by line, without any additional content.

[Answer]
MethodA(java.lang.String)
MethodB()
MethodC(int, java.lang.String)
\end{lstlisting}
\end{promptbox}

\subsubsection{T3-h: Localized Interface Change Detection}

Detect local differences between two versions of an interface or resource model.

\begin{promptbox}[T3-h: Localized Interface Change Detection]
\begin{lstlisting}[style=promptstyle]
Primary Task: Version and Code Diff Analysis
Compare changes in different text/code versions.

Secondary Task: Localized Interface Change Detection
Detect local version differences.
Context Requirement: Partial
Metric: F1

Instruction:
Identify fields that were renamed, removed, or refactored between versions. Use the format "Model: [Field Name]".

Output format:
Output the "[Answer]" identifier first, then output fields line by line, without any additional content.

[Answer]
Lesson Model: key
Lesson Model: seq length
\end{lstlisting}
\end{promptbox}

% =========================================================
%  T4 -- Numerical Extraction and Quantitative Reasoning
% =========================================================
\subsection{T4: Numerical Extraction and Quantitative Reasoning}
\label{app:prompt_t4}

T4 evaluates the model's ability to perform numerical calculations within structured text such as financial tables and reports. The reward function uses \texttt{math\_verify}. Subtask prompts cover both multi-source consistency verification and single-source targeted aggregation.

\subsubsection{T4-a: Multi-Source Consistency Verification}

Verify numerical consistency across multiple structured sources using a given formula.

\begin{promptbox}[T4-a: Multi-Source Consistency Verification]
\begin{lstlisting}[style=promptstyle]
Primary Task: Structured and Numeric Reasoning
Numerical calculations in structured text.

Secondary Task: Structured Multi-Source Consistency Verification
Numerical computation across multiple sources.
Context Requirement: Full
Metric: math_verify

Instruction:
Given multiple tables, use the provided formula to determine whether the computed values are consistent across sources. If consistent, return "No Error". If inconsistent, return the corresponding record IDs.

Output format:
Output the "[Answer]" identifier first, then output "No Error" or the corresponding IDs line by line, without any additional content.

[Answer]
No Error
\end{lstlisting}
\end{promptbox}

\subsubsection{T4-b: Single-Source Targeted Aggregation}

Perform targeted computations within a single structured source.

\begin{promptbox}[T4-b: Single-Source Targeted Aggregation]
\begin{lstlisting}[style=promptstyle]
Primary Task: Structured and Numeric Reasoning
Numerical calculations in structured text.

Secondary Task: Single-Source Targeted Aggregation
Query computation in a single source.
Context Requirement: Partial
Metric: math_verify

Instruction:
According to the specified table, calculate the requested quantities (e.g., percentage changes).

Output format:
Output the "[Answer]" identifier first, then output the computed values line by line, without any additional content.

[Answer]
5.5%
-2.2%
\end{lstlisting}
\end{promptbox}

\subsubsection{T4-c: Procedural State Tracking}

Track entity state evolution across a long procedural narrative.

\begin{promptbox}[T4-c: Procedural State Tracking]
\begin{lstlisting}[style=promptstyle]
Primary Task: Structured and Numeric Reasoning
Numerical calculations in structured text.

Secondary Task: Long-Context Procedural State Tracking
Track entity state evolution across the full context.
Context Requirement: Full
Metric: math_verify

Output format:
Output the "[Answer]" identifier first, then output tracked states line by line, without any additional content.

[Answer]
state_1
state_2
\end{lstlisting}
\end{promptbox}

% =========================================================
%  T5 -- Multi-Table Structured Extraction (Code Execution)
% =========================================================
\subsection{T5: Multi-Table Structured Extraction}
\label{app:prompt_t5}

T5 evaluates the model's ability to extract and integrate information from multiple structured tables. The reward function uses IoU-based Structured Match, which deserializes both the predicted JSON and the reference JSON into structured objects and computes the Intersection-over-Union of their attributes. Deserialization eliminates the effect of attribute ordering, so the comparison is based solely on structural and value-level overlap. Training data for this task is curated from open-source multi-table QA corpora and does not require additional generation prompts beyond the shared instruction.

\begin{promptbox}[T5: Multi-Table Structured Extraction]
\begin{lstlisting}[style=promptstyle]
Primary Task: Structured Table Reasoning
Perform multi-table querying, aggregation, filtering, and relational reasoning.

Secondary Task: Long-Context Multi-Table Integration
Integrate information distributed across multiple tables within the full context.
Context Requirement: Full
Metric: IoU-based Structured Match

Instruction:
You are given multiple tables and a question. Carefully read all tables before reasoning. When combining tables, explicitly identify the join keys and referenced columns. If aggregation, filtering, or sorting is required, reason step by step before producing the answer.

Output format:
Output the final answer strictly between <answer> and </answer> tags. The answer must be a valid JSON object: {"columns": ["col1", ...], "data": [["val1", ...], ...]}

If the answer cannot be inferred from the provided tables, output: <answer>{"columns": [], "data": []}</answer>

[Answer]
formatted output
\end{lstlisting}
\end{promptbox}
% =========================================================
%  T6 -- Fragment-Level Structured Matching and Induction (SubEM)
% =========================================================
\subsection{T6: Fragment-Level Structured Matching and Induction}
\label{app:prompt_t6}

T6 assesses the model's ability to perform clustering, rule induction, and structured matching at the fragment level. The reward function is Substring Exact Match (SubEM). This task covers large-scale document clustering, rule induction from examples, and related pattern recognition subtasks.

\subsubsection{T6-a: Large-Scale Document Clustering}

Cluster documents by a specified criterion and report category proportions.

\begin{promptbox}[T6-a: Large-Scale Document Clustering]
\begin{lstlisting}[style=promptstyle]
Primary Task: Aggregation and Clustering
Cluster and output statistics, examples, or sorted results.

Secondary Task: Large-Scale Document Clustering
Return all category proportions.
Context Requirement: Full
Metric: SubEM

Instruction:
Cluster the given documents by the specified criterion into the required number of clusters.

Output format:
Output the "[Answer]" identifier first, then output the clusters in the format "ClusterName Proportion%" (rounded to two decimal places) line by line, without any additional content.

[Answer]
Cluster_A 25.00%
Cluster_B 25.00%
Cluster_C 50.00%
\end{lstlisting}
\end{promptbox}

\subsubsection{T6-b: Large-Scale In-Context Rule Induction}

Induce formatting or transformation rules from global context examples and apply them to new input.

\begin{promptbox}[T6-b: In-Context Rule Induction]
\begin{lstlisting}[style=promptstyle]
Primary Task: Rule Induction and In-Context Learning
Summarize rules and make decisions on new samples.

Secondary Task: Large-Scale In-Context Rule Induction
Induce rules from the global context.
Context Requirement: Full
Metric: SubEM

Instruction:
Based on the conventions demonstrated in the provided text, apply the induced rules to reformat or transform the given input snippet.

Output format:
Output the "[Answer]" identifier first, then output the fully formatted result, without any additional content.

[Answer]
formatted output
\end{lstlisting}
\end{promptbox}

\subsubsection{T6-c: Targeted Example-Based Rule Induction}

Induce rules from targeted examples and apply them to a new case for decision-making.

\begin{promptbox}[T6-c: Targeted Rule Induction]
\begin{lstlisting}[style=promptstyle]
Primary Task: Rule Induction and In-Context Learning
Summarize rules and make decisions on new samples.

Secondary Task: Targeted Example-Based Rule Induction
Induce rules from the targeted examples.
Context Requirement: Partial
Metric: SubEM

Instruction:
Based on the case examples in the specified section, answer the given question by applying the induced rule. The answer must be a fixed value.

Output format:
Output the "[Answer]" identifier first, then output the answer, without any additional content.

[Answer]
support
\end{lstlisting}
\end{promptbox}

% =========================================================
%  T7 -- Dimension-Quantified Retrieval and Graded Ranking (NDCG)
% =========================================================
\subsection{T7: Dimension-Quantified Retrieval and Graded Ranking}
\label{app:prompt_t7}

T7 evaluates the model's ability to retrieve content and rank results by relevance or a quantified dimension. The reward function is NDCG.

\subsubsection{T7-a: Global Cohesive Retrieval}

Retrieve all qualifying items from the full text and output them in the specified order.

\begin{promptbox}[T7-a: Global Cohesive Retrieval]
\begin{lstlisting}[style=promptstyle]
Primary Task: Retrieval and Ranking
Retrieve content and rank most relevant first.

Secondary Task: Global Cohesive Retrieval
Retrieve from the full text and reorganize.
Context Requirement: Full
Metric: NDCG@k

Instruction:
Retrieve all items that meet the specified condition and output their IDs in the required sorted order.

Output format:
Output the "[Answer]" identifier first, then output IDs line by line, without any additional content.

[Answer]
A2AV7Q95QGPTO0
A3NM0RAYSL6PA8
A1C9C1QOQB94RT
\end{lstlisting}
\end{promptbox}

\subsubsection{T7-b: Key-Snippet Retrieval}

Locate and rank target fragments within a specified subset of the document.

\begin{promptbox}[T7-b: Key-Snippet Retrieval]
\begin{lstlisting}[style=promptstyle]
Primary Task: Retrieval and Ranking
Retrieve content and rank most relevant first.

Secondary Task: Key-Snippet Retrieval
Locate target fragment in a specified paragraph subset.
Context Requirement: Partial
Metric: NDCG@k

Instruction:
From the specified subset, retrieve items matching the given category and sort them by the requested criterion.

Output format:
Output the "[Answer]" identifier first, then output the sorted IDs line by line, without any additional content.

[Answer]
ID1
ID2
ID3
\end{lstlisting}
\end{promptbox}

% =========================================================
%  T8 -- Sequence Reconstruction and Ordering (Pairwise)
% =========================================================
\subsection{T8: Sequence Reconstruction and Ordering}
\label{app:prompt_t8}

T8 evaluates the model's ability to restore temporal, logical, or frequency-based orderings. The reward function is Pairwise Accuracy.

\subsubsection{T8-a: Global Timeline Reconstruction}

Sort shuffled segments of the full text into their correct chronological order.

\begin{promptbox}[T8-a: Global Timeline Reconstruction]
\begin{lstlisting}[style=promptstyle]
Primary Task: Sequencing and Structure Reconstruction
Restore timeline or logical order.

Secondary Task: Global Timeline Reconstruction
Sort unordered events across the whole text.
Context Requirement: Full
Metric: Pairwise Accuracy

Instruction:
The text is divided into multiple parts in shuffled order. Sort these parts in the correct chronological sequence.

Output format:
Output the "[Answer]" identifier first, then output the correct sequence of part identifiers line by line, without any additional content.

[Answer]
Part 5
Part 1
Part 9
Part 3
\end{lstlisting}
\end{promptbox}

\subsubsection{T8-b: Local Causal Chain Sorting}

Reorder content within a specific paragraph according to causal or logical structure.

\begin{promptbox}[T8-b: Local Causal Chain Sorting]
\begin{lstlisting}[style=promptstyle]
Primary Task: Sequencing and Structure Reconstruction
Restore timeline or logical order.

Secondary Task: Local Causal Chain Sorting
Sort content in a specific paragraph.
Context Requirement: Partial
Metric: Pairwise Accuracy

Instruction:
The specified paragraph is out of order. Reorder it according to the original text and options.

Output format:
Output the "[Answer]" identifier first, then output the sorted option letters line by line, without any additional content.

[Answer]
A
B
C
D
E
\end{lstlisting}
\end{promptbox}

\subsubsection{T8-c: Global Frequency Analysis}

Count and sort terms by their frequency of occurrence across the full text.

\begin{promptbox}[T8-c: Global Frequency Analysis]
\begin{lstlisting}[style=promptstyle]
Primary Task: Aggregation and Clustering
Cluster and output statistics, examples, or sorted results.

Secondary Task: Global Frequency Analysis
Count and sort global word frequency.
Context Requirement: Full
Metric: Pairwise Accuracy

Instruction:
Sort the given terms in descending order by their frequency of appearance in the text.

Output format:
Output the "[Answer]" identifier first, then output the sorted terms line by line, without any additional content.

[Answer]
would
this
that
went
have
\end{lstlisting}
\end{promptbox}

% =========================================================
%  T9 -- Long Document Summarization (Summary)
% =========================================================
\subsection{T9: Long Document Summarization}
\label{app:prompt_t9}

T9 evaluates the model's ability to generate abstractive summaries under specified constraints. The reward function is ROUGE-L.

\subsubsection{T9-a: Global-Coverage Constrained Summary}

Summarize the full document under a specified word limit.

\begin{promptbox}[T9-a: Global-Coverage Constrained Summary]
\begin{lstlisting}[style=promptstyle]
Primary Task: Summarization and Synthesis
Generate abstract summary under given constraints.

Secondary Task: Global-Coverage Constrained Summary
Generate summary of the full text.
Context Requirement: Full
Metric: ROUGE-L

Output format:
Output the "[Answer]" identifier first, then output the summary, without any additional content.

[Answer]
your summary
\end{lstlisting}
\end{promptbox}

\subsubsection{T9-b: Query-Focused Summary}

Summarize a specific subtopic or section within the document.

\begin{promptbox}[T9-b: Query-Focused Summary]
\begin{lstlisting}[style=promptstyle]
Primary Task: Summarization and Synthesis
Generate abstract summary under given constraints.

Secondary Task: Query-Focused Summary
Generate summary of a specific subtopic.
Context Requirement: Partial
Metric: ROUGE-L

Output format:
Output the "[Answer]" identifier first, then output the summary, without any additional content.

[Answer]
your summary
\end{lstlisting}
\end{promptbox}

\section{Evaluation Alignment with QwenLong-L1.5}
% \subsection{Evaluation Alignment with QwenLong-L1.5}
\label{app:eval_alignment}

% To ensure a fair comparison with QwenLong-L1.5, we verify that our evaluation pipeline faithfully reproduces the results reported in the original paper. Tables~\ref{tab:eval_alignment_lc},~\ref{tab:eval_alignment_general}, and~\ref{tab:eval_alignment_memory} present the reported scores alongside our reproduced results using the same released checkpoints. The difference ($\Delta$) is shown as a colored subscript next to each reproduced score (\textcolor{green!50!black}{green} for positive, \textcolor{red!70!black}{red} for negative). Across most benchmarks, our reproduced scores closely match the reported values, confirming that our evaluation protocol is well-aligned with that of QwenLong-L1.5.

To ensure a fair comparison with QwenLong-L1.5, we verify that our evaluation pipeline faithfully reproduces the results reported in the original paper. Tables~\ref{tab:eval_alignment_lc},~\ref{tab:eval_alignment_general}, and~\ref{tab:eval_alignment_memory} present the published scores with $\Delta = \text{ours} - \text{published}$ shown as a subscript next to each value. Deviations within a small margin are shown in {\color{gray}gray}; larger deviations are highlighted in {\color{orange!85!black}orange}. Across most benchmarks, the deviations are small, confirming that our evaluation protocol is well-aligned with that of QwenLong-L1.5.

\newcommand{\dltg}[1]{{\color{gray!70}\scriptsize$_{\,#1}$}}          % 小偏差
\newcommand{\dltw}[1]{{\color{orange!85!black}\bfseries\scriptsize$_{\,#1}$}}  % 大偏差

\begin{table}[h]
\centering
\small
\setlength{\tabcolsep}{6pt}
\renewcommand{\arraystretch}{1.35}
\caption{Evaluation protocol alignment against the QwenLong-L1.5 paper.
Each cell reports the published score with $\Delta = \text{ours} - \text{published}$
as a subscript.}
\label{tab:eval_alignment_lc}
\begin{tabular}{lcccccc}
\toprule
\multirow{2}{*}{\textbf{Model}}
  & \textbf{DocMath} & \textbf{LBV2} & \textbf{Frames}
  & \textbf{MRCR}   & \textbf{CorpusQA} & \textbf{LBV1-QA} \\
& \multicolumn{6}{c}{\footnotesize\textit{(published score, $\Delta$= ours $-$ published)}} \\
\midrule
Qwen3-4B-Thinking
  & $59.0$\dltg{+2.0}
  & $41.4$\dltg{-1.2}
  & $62.9$\dltg{+1.5}
  & $39.9$\dltg{-1.5}
  & $49.9$\dltg{\pm0.0}
  & $64.3$\dltg{-0.3} \\
\midrule
Qwen3-30B-A3B-Thinking
  & $62.3$\dltg{+1.0}
  & $49.1$\dltg{-0.4}
  & $70.3$\dltg{-0.1}
  & $51.3$\dltw{-9.7}
  & $71.6$\dltg{-1.1}
  & $67.1$\dltg{-0.6} \\
QwenLong-L1.5-30B-A3B
  & $66.3$\dltg{+0.1}
  & $55.3$\dltg{-0.1}
  & $74.8$\dltg{-0.3}
  & $83.0$\dltg{-0.5}
  & $81.3$\dltg{-0.4}
  & $70.4$\dltg{-2.7} \\
\bottomrule
\end{tabular}
\end{table}

\begin{table}[b]
\centering
\small
\setlength{\tabcolsep}{7pt}
\renewcommand{\arraystretch}{1.35}
\caption{Evaluation protocol alignment on general reasoning benchmarks.
Each cell reports the published score with $\Delta = \text{ours} - \text{published}$
as a subscript.}
\label{tab:eval_alignment_general}
\begin{tabular}{lcccc}
\toprule
\multirow{2}{*}{\textbf{Model}}
  & \textbf{MMLU-Pro} & \textbf{AIME24} & \textbf{AIME25} & \textbf{GPQA} \\
& \multicolumn{4}{c}{\footnotesize\textit{(published score,
  $\Delta$ = ours $-$ published)}} \\
\midrule
Qwen3-30B-A3B-Thinking
  & $81.0$\dltg{-0.8}
  & $90.3$\dltg{+0.2}
  & $82.8$\dltg{+2.3}
  & $75.9$\dltw{-5.8} \\
QwenLong-L1.5-30B-A3B
  & $81.3$\dltg{-0.2}
  & $90.0$\dltg{-0.2}
  & $86.5$\dltg{+1.4}
  & $76.8$\dltw{-4.2} \\
\bottomrule
\end{tabular}
\end{table}

% \begin{table}[h]
% \centering
% \caption{Evaluation alignment on agentic memory and dialogue memory benchmarks.}
% \label{tab:eval_alignment_memory}
% \resizebox{\linewidth}{!}{%
% \renewcommand{\arraystretch}{1.3}
% \begin{tabular}{l*{4}{cc}}
% \toprule
% & \multicolumn{2}{c}{\textbf{Memory-KV}} & \multicolumn{2}{c}{\textbf{Memory-Vec}} & \multicolumn{2}{c}{\textbf{Mem-Rec\_Sum}} & \multicolumn{2}{c}{\textbf{LongMemEval}} \\
% \cmidrule(lr){2-3} \cmidrule(lr){4-5} \cmidrule(lr){6-7} \cmidrule(lr){8-9}
% & Rep. & Ours & Rep. & Ours & Rep. & Ours & Rep. & Ours \\
% \midrule
% Qwen3-30B-A3B-Thinking & 11 & 15.48\posdelta{4.5} & 16.1 & 18.71\posdelta{2.6} & 41.9 & 30.97\negdelta{10.9} & 60.8 & 60.8\zerodelta \\
% QwenLong-L1.5-30B-A3B & 16.8 & 16.13\negdelta{0.7} & 16.8 & 8.1\negdelta{8.7} & 40 & 36.13\negdelta{3.9} & 76.4 & 72.2\negdelta{4.2} \\
% \bottomrule
% \end{tabular}%
% }
% \end{table}

\begin{table}[h]
\centering
\small
\setlength{\tabcolsep}{7pt}
\renewcommand{\arraystretch}{1.35}
\caption{Evaluation protocol alignment on agentic and dialogue memory benchmarks.
Each cell reports the published score with $\Delta = \text{ours} - \text{published}$
as a subscript.}
\label{tab:eval_alignment_memory}
\begin{tabular}{lcccc}
\toprule
\multirow{2}{*}{\textbf{Model}}
  & \textbf{Memory-KV} & \textbf{Memory-Vec} & \textbf{Mem-Rec\_Sum} & \textbf{LongMemEval} \\
& \multicolumn{4}{c}{\footnotesize\textit{(published score,
  $\Delta$ = ours $-$ published)}} \\
\midrule
Qwen3-30B-A3B-Thinking
  & $11.0$\dltw{+4.5}
  & $16.1$\dltg{+2.6}
  & $41.9$\dltw{-10.9}
  & $60.8$\dltg{\pm0.0} \\
QwenLong-L1.5-30B-A3B
  & $16.8$\dltg{-0.7}
  & $16.8$\dltw{+4.5}
  & $40.0$\dltg{-3.9}
  & $76.4$\dltg{-4.2} \\
\bottomrule
\end{tabular}
\end{table}

% =============================================================
%  Appendix C -- Experimental Details
% =============================================================
\section{Training Hyperparameters}
\label{app:train_config}

Table~\ref{tab:hyperparams} summarizes the key hyperparameters used in our RL training. All configurations are shared across the two model scales.

\begin{table}[t!]
\centering
\small
\caption{Training hyperparameters for GoLongRL.}
\label{tab:hyperparams}
\setlength{\tabcolsep}{10pt}
\renewcommand{\arraystretch}{1.15}
\begin{tabularx}{\linewidth}{@{}Xr@{}}
\toprule
\textbf{Hyperparameter} & \textbf{Value} \\
\midrule
\multicolumn{2}{@{}l}{\textit{Data}} \\
\quad Max prompt length & 160K \\
\quad Max response length & 16K \\
\quad Responses per prompt & 16 \\
\midrule
\multicolumn{2}{@{}l}{\textit{Optimization}} \\
\quad Learning rate & 2e-6 \\
\quad LR warmup steps & 5 \\
\quad Weight decay & 0.1 \\
\quad Gradient clipping & 1.0 \\
\quad PPO epochs & 1 \\
\quad Total epochs & 10 \\
\quad Train prompt batch size & 128 \\
\quad Loss aggregation & token-mean \\
\midrule
\multicolumn{2}{@{}l}{\textit{Clipping}} \\
\quad Clip ratio (low) & 0.2 \\
\quad Clip ratio (high) & 0.28 \\
\quad Clip ratio $c$ & 3.0 \\
\midrule
\multicolumn{2}{@{}l}{\textit{Importance Sampling}} \\
\quad IS level & token \\
\quad IS clipping mode & clip \\
\quad IS threshold (upper) & 5.0 \\
\quad IS threshold (lower) & 0.5 \\
\quad IS veto threshold & 1e-4 \\
\midrule
\multicolumn{2}{@{}l}{\textit{TMN-Reweight}} \\
\quad Difficulty reweighting & enabled \\
\quad Smoothing $\alpha$ & 0.8 \\
\bottomrule
\end{tabularx}
\end{table}

% \section{Additional Experimental Details}
% \label{app:experiments}

% % Placeholder for experimental appendix content.
% \textcolor{red}{\textbf{[TODO]} This section will include additional experimental details, including:
% \begin{itemize}[nosep]
%     \item Detailed hyperparameter configurations for all training runs.
%     \item Full evaluation protocol and benchmark reproduction details.
%     \item Generalization results on general reasoning benchmarks (MMLU-Pro, AIME24/25, GPQA-Diamond).
%     \item Agentic memory and dialogue memory evaluation results (BFCL-V4, LongMemEval).
%     \item Extended ablation studies and training curve analysis.
% \end{itemize}}